\documentclass[sn-basic]{sn-jnl}% Math and Physical Sciences Reference Style
\jyear{2023}%

\usepackage[english]{babel}

\usepackage[utf8]{inputenc}
\usepackage{graphicx, setspace}

\usepackage{amsmath}
\usepackage{amssymb}
\usepackage{amsthm}
\usepackage{xcolor}
\usepackage{multirow}
\usepackage{comment}
\usepackage{todonotes}

\theoremstyle{thmstyleone}%
\newcommand{\mB}{\mathcal{B}}

\theoremstyle{thmstyletwo}%

\newtheorem{lemma}{Lemma}%

\theoremstyle{thmstylethree}%

\begin{document}

\title{Fast Neighborhood Search Heuristics for the Colored Bin Packing Problem}

\author*[1]{\fnm{Renan F.} \sur{F. da Silva}}\email{renan.silva@students.ic.unicamp.br}

\author[1]{\fnm{Yulle G.} \sur{F. Borges}}\email{glebbyo@ic.unicamp.br}

\author[1]{\fnm{Rafael C.} \sur{S. Schouery}}\email{rafael@ic.unicamp.br}

\affil[1]{\orgdiv{Institute of Computing}, \orgname{University of Campinas}, \orgaddress{\city{Campinas}, \country{Brazil}}}

\abstract{The Colored Bin Packing Problem (CBPP) is a generalization of the Bin Packing Problem~(BPP). The CBPP consists of packing a set of items, each with a weight and a color, in bins of limited capacity, minimizing the number of used bins and satisfying the constraint that two items of the same color cannot be packed side by side in the same bin. In this article, we proposed an adaptation of BPP heuristics and new heuristics for the CBPP\@. Moreover, we propose a set of fast neighborhood search algorithms for CBPP\@. These neighborhoods are applied in a meta-heuristic approach based on the Variable Neighborhood Search~(VNS) and a matheuristic approach that combines linear programming with the meta-heuristics VNS and Greedy Randomized Adaptive Search~(GRASP). The results indicate that our matheuristic is superior to VNS and that both approaches can find near-optimal solutions for a large number of instances, even for those with many items.}


\maketitle

\section{Introduction}

In the Bin Packing Problem (BPP), there is a set of items and an unlimited set of bins of equal capacity. The objective is to allocate the items in bins, respecting their capacities and minimizing the number of used bins. A generalization of BPP is the Colored Bin Packing Problem (CBPP), where each item also has a color, and an additional constraint is imposed: items of the same color cannot be placed adjacent to each other within the same bin.

The CBPP was introduced by~\cite{Dosa_2014}, where they primarily focused on online and approximation algorithms for solving CBPP\@. A year before this contribution,~\cite{Balogh_2012} introduced the Black and White Bin Packing Problem (BWBPP), a particular case of CBPP with only two colors.

\cite{Balogh_2012} also identified two practical applications for CBPP\@. One of which involves packing two-dimensional objects with similar widths in an alternating manner. An illustrative example is the layout of information and advertisements on platforms like YouTube and social media, where content must be stacked to fit within the limited screen space of smartphones. Here, the bin's capacity corresponds to the maximum allowable content for one screen, the height of each piece of content represents its size, informational content is designated as white, and advertisements are designated as black.

The second application pertains to scheduling commercial breaks on radio or television, with varying lengths and genres, which must be organized into commercial blocks. Here, the block's duration corresponds to the bin capacity, while the duration of each commercial serves as the item size. To ensure diversity within the breaks, commercials of the same genre are assigned the same color, preventing them from being aired consecutively.

The CBPP is an NP-hard problem since it is a generalization of BPP~\citep{Karp_1972}. In tackling NP-hard problems like CBPP, besides exact algorithms, heuristic algorithms offer another avenue of exploration, particularly through the meta-heuristic approach. Meta-heuristics are renowned for producing high-quality solutions within reasonable computational time.

This article delves into applying meta-heuristics and matheuristic to address CBPP, representing, to the best of our knowledge, the first study in the literature to use these approaches. Next, we introduce a literature review of heuristics for BPP and present our contributions to CBPP\@.

\subsection{Literature Review}

In the online version of CBPP, items are received sequentially and must be permanently allocated into a chosen bin, without prior knowledge of subsequent arrivals. Conversely, the offline version operates with full knowledge of the input. We categorize the offline version as either unrestricted or restricted depending on whether the items can or not be reordered before the packing process starts. An online algorithm is deemed~$\alpha$-competitive if, for every instance~$I$, its solution value is at most~$\alpha$ times~$OPT(I)$, where~$OPT(I)$ is the optimal solution value for~$I$ in the offline restricted version.

\cite{Balogh_2012} introduced the BWBPP\@. They established a lower bound of 1 +~$\frac{1}{
2 \cdot \ln 2} \approx 1.7213$ for the asymptotic competitive ratio of any online algorithm and presented an absolute~$3$-competitive algorithm for the online version. Additionally, they proposed a~$2.5$-approximation algorithm and an Asymptotic Polynomial-Time Approximation Scheme (APTAS) for the offline unrestricted version.
These findings were subsequently published in journal papers~\citep{Balogh_2015, Balogh_2015b}, which also included extensions. Particularly, \cite{Balogh_2015} also demonstrated how to modify this APTAS to yield both an Asymptotic Fully Polynomial-Time Approximation Scheme and an absolute~$\frac{3}{2}$-approximation algorithm, which is the best possible approximation ratio, unless P = NP\@.

\cite{Dosa_2014} demonstrated that no optimal online algorithm exists for the zero-sized items variant of the CBPP, contrasting with BWBPP where such an algorithm exists. Furthermore, they introduced a 4-competitive online algorithm and a lower bound of 2 on the asymptotic competitive ratio of any online algorithm. Given that the BWBPP is a particular case of the CBPP, this latter result enhances the lower bound initially proposed by~\cite{Balogh_2012}.

In a conference paper \citep{Bohm_2014}, later published as full paper~\citep{Bohm_2018}, Böhm et al.\ revealed that adaptations of the classic BPP algorithms --- First Fit, Best Fit, and Worst Fit --- for the online CBPP do not have constant competitive ratios. Additionally, they refined the findings of~\cite{Dosa_2014}, demonstrating that any online algorithm has an asymptotic competitive ratio of at least 2.5 and introducing a (absolute) 3.5-competitive algorithm.

\cite{Chen_2015} delved into the special case of the BWBPP, where items have lengths of at most half the capacity of the bin, and introduced a~$\frac{8}{3}$-competitive algorithm for it.
This algorithm outperforms the one presented by~\cite{Balogh_2012}, which maintains a competitive ratio of 3 in this scenario.

\cite{Alsarhan_2016} proposed an exact and linear time algorithm for the offline unrestricted variants of the CBPP, in which all items have zero or unitary weights. Subsequently,~\cite{Borges_2023} introduced five exact algorithms for the CBPP\@. The first algorithm adapts the classic compact BPP model introduced by~\cite{Martello1990} for formal purposes (a model previously misattributed to~\cite{Kantorovich1939}, as revealed by~\cite{Uchoa2024}). The following two algorithms present pseudo-polynomial formulations based on the Arc Flow Formulation for the BPP proposed by~\cite{Carvalho_1999}. Finally, the last two algorithms employ set-partition formulations and are solved using the VRPSolver framework proposed by~\cite{Pessoa_2020}.

Numerous authors have extensively explored the BPP and its variants in the literature, employing both exact algorithms and meta-heuristic approaches. Here, we present an overview of the most prominent algorithms for the BPP and its generalization called the Cutting Stock Problem (CSP). Additionally, we highlight key studies employing VNS in the context of the BPP\@.

Exact algorithms for the BPP primarily rely on two main formulation types: the Kantorovich-Gilmore-Gomory Formulation, initially proposed by~\cite{Kantorovich1951} and \cite{Gilmore_1961} (refer to \cite{Uchoa2024}, for an understanding of this renaming), and the Arc Flow Formulation introduced by~\cite{Carvalho_1999}. Works utilizing the Kantorovich-Gilmore-Gomory formulation include those by~\cite{Vance_1998},~\cite{Belov_2006}, and~\cite{Wei_2019}. Conversely, studies employing the Arc Flow Formulation include those by~\cite{Brandao_2016},~\cite{Delorme_2020}, and~\cite{Loti_2022}.

On the other hand, in the realm of heuristic algorithms,~\cite{Fleszar_2002} delve into the application of the Variable Neighborhood Search~(VNS) meta-heuristic to BPP\@. They adopt a variation of the heuristic known as Minimum Bin Slack, initially proposed by~\cite{Gupta_1999}, as an initial solution for VNS\@. Their computational experimentation employs a benchmark comprising Faulkenauer's and Scholl's instances, available in the BPP Lib~\citep{Delorme_2018}. The MBS variant integrated into the proposed VNS demonstrates significant efficacy, achieving optimal solutions in 1329 of the 1370 instances examined.

Subsequently,~\cite{Loh_2008} implement the Weight Annealing meta-heuristic in the BPP, achieving optimal solutions in all instances used by Fleszar and Hindi.~\cite{Castelli_2014} introduce a BPP heuristic named BPHS, utilized as the initial solution of the VNS\@. Unfortunately, this combination exhibited a lower performance compared to the two preceding studies.

Finally,~\cite{Gonzalez_2023} present a comparative study of nineteen heuristic algorithms proposed in the last two decades. The top-performing algorithm in the study was Consistent Neighborhood Search~(CNS), introduced by~\cite{Buljubasic_2016}, which archives an optimal solution in 99.81\% of the studied instances.

\subsection{Our Contributions}
In this article, we investigate the CBPP utilizing five initial construction heuristics. The first two are adaptations of heuristics of the BPP known as Best Fit Decreasing and Minimum Bin Slack. The remaining three are heuristics developed by us, termed Good Ordering, Hard BFD, and Two-by-Two. Later, we introduce a meta-heuristic approach for CBPP employing the VNS, which leverages the aforementioned heuristics to construct the initial solution. Finally, we propose a new matheuristic that integrates linear programming with the meta-heuristics VNS and Greedy Randomized Adaptive Search~(GRASP). Our findings demonstrate that our approaches can achieve near-optimal solutions for instances containing up to~$10^4$ items within seconds. Particularly, for all instances studied comprising up to~$2 \times 10^3$ items and possessing known optimal solutions, our approaches, within 60 seconds, produce solutions utilizing at most one additional bin compared to the optimal solution.

The article is structured as follows. Section~\ref{section:CBPP} provides a formal definition of the CBPP\@. Subsequently, in Section~\ref{section:Notations}, we introduce convenient notations and establish a proposition that simplifies the handling of the color constraint. Section~\ref{section:Auxiliar} presents an Auxiliary Algorithm utilized as a subroutine in some of our algorithms. The initial heuristics, the VNS, and the matheuristic are detailed, respectively, in Sections~\ref{section:Initial},~\ref{section:VNS}, and~\ref{section:MH}. Finally, Sections~\ref{section:Results} and~\ref{section:Conclusion} contain the computational results and conclusions.

\section{Colored Bin Packing Problem}\label{section:CBPP}

In the CBPP, we are given a set of bins with capacity W and a set~${I = \{1, \dots, n \}}$ of items, where each item~$i$ has a weight~$0 < w_i \leq W$ and a color~$c_i \in \mathbb{N}$. The objective is to pack each item~$i \in I$ into the bins~$\{B_1, \dots, B_k\}$, minimizing~$k$ while satisfying the capacity and color constraints. Specifically, for every~$1 \leq r \leq k$, we must satisfy the capacity constraint~$\sum_ {i \in B_r} w_i \leq W$ and the color constraint: there exists a permutation~$\pi~$ of~$B_r$ such that, for every item pair~$(i, j)$ in~$B_r$, if~$c_i = c_j$, then~$\pi (i) \neq \pi (j) + 1$, in other words, two items with the same color cannot be packed side by side.

A subset~$p \subseteq I$ of items is termed a \textit{pattern} if it satisfies the capacity and color constraints. A solution for the CBPP comprises a set of patterns where each item belongs to precisely one pattern. Next, we present a possible formulation for the CBPP, which is based on the Kantorovich-Gilmore-Gomory Formulation proposed by~\cite{Kantorovich1951} and~\cite{Gilmore_1961} for the BPP\@:
\begin{alignat}{3}
 \mathrm{(P)}\quad& minimize  \quad  && \displaystyle{\sum_{p \in \mathcal{P}} \lambda_p}\nonumber\\
              & subject~to \quad       &&\displaystyle{\sum_{p \in \mathcal{P}\wedge i \in p} \lambda_p = 1,}    \quad&& i \in I\\
              & \quad       &&\lambda_p \in \{0, 1\}, \quad    && \forall p \in \mathcal{P}
\end{alignat}
where~$\mathcal{P}$ is the set of all patterns, and for each~$p \in \mathcal{P}$, there is a binary variable~$\lambda_p$ that is equal to 1 if, and only if, the pattern~$p$ is used. The objective function seeks to minimize the number of patterns used. It is worth noting that the cardinality of~$\mathcal{P}$ cannot be polynomially bounded by the number of items. Consequently, this formulation is usually solved by column generation, just like done by~\cite{Belov_2006}, \citet{Wei_2019} and~\citet{Baldacci_2024} for the BPP\@.

\section{Notations\label{section:Notations}}

Considering a fixed feasible solution for the CBPP, we employ throughout this article the following notations: item~$i$ is packed in bin~$B(i)$; the residual capacity~$res(r)$ of bin~$B_r$ is~$W - \sum _ {j \in B_r} w_j$; and~$tight(r)$ is the color that if added in~$B_r$ violates the color constraint. If this color does not exist, we define~$tight(r) = \emptyset$. As we will see, if any color satisfies this property, then this color is unique.

Furthermore, we assume that the items are sorted by non-increasing weight, and the bins of a solution are sorted by residual capacity. Whether these sorting orders are non-increasing or non-decreasing depends on the algorithms utilized. We adopt this approach because these algorithms are invoked multiple times, using, as input, the output of the previous call. Given that these algorithms manipulate only a constant number of items, we can maintain both sets ordered with a maximum of~$O(n)$ computational steps.

In addition,~\cite{Borges_2023} presents a necessary and sufficient condition for the existence of a permutation~$\pi$ of bin~$B_r$ that satisfies the color constraint without the need to build it. Lemma 1 introduces this condition using the notation employed in this paper, and its proof is provided by~\cite{Borges_2023}. Due to the simplicity of this condition, our algorithms primarily focus on ensuring its fulfillment. Thus, given the final solution~$(B_1, \ldots, B_k)$, we can have a post-processing that builds~$\pi$ for each~$B_r$ with time complexity~$\Theta( \vert B_r \vert  \log  \vert B_r \vert )$ using the algorithm proposed by~\cite{Borges_2023}. By performing this operation for each bin~$1 \leq r \leq k$, we can obtain the permutation for all bins in the final solution within time~$O(n \log n)$.

\begin{lemma}
 Given a subset~$B_r$ of items. Let~$g$ be one of the most frequent colors of~$B_r$,~$B_r^g$ be the subset of~$B_r$ of items with color~$g$, and~$B_r^o$ be the subset of~$B_r$ of items with a color different from~$g$. Then, there is a permutation~$\pi$ of~$B_r$ that satisfies color constraint if, and only if,~$ \vert B_r^g \vert  \leq \vert B_r^o \vert  + 1$. Also, if~$ \vert B_r^g \vert  =  \vert B_r^o \vert  + 1$, then~$tight(r) = g$ and~$g$ is unique.
\end{lemma}

\section{Auxiliary Algorithm}\label{section:Auxiliar}

Given a set~$I$ of items sorted in non-increasing order of weight and a set of bins~$\mB$ sorted in non-increasing order of residual capacity, several of our algorithms must find, for each~$i\in I$, the index of the fullest bin~$B' \in \mB$ where~$i$ can be packed.  In certain scenarios,~$B(i) \neq \emptyset$, which indicates that~$i$ is already packed in a bin of~$\mB$; in such cases, we stipulate that~$B' \neq B(i)$. This requirement is crucial for our neighborhoods presented later, which seek to relocate the item~$i$ to another bin. For simplicity, if an item~$i \in I$ lacks a bin~$B'$ meeting these constraints, we assign the answer for item~$i$ as 0 (an invalid index).

Next, we show an algorithm with time complexity~$\Theta( \vert I \vert  +  \vert \mB \vert )$ to solve this problem. The algorithm aims to find a set of pairs~$\{(i, t(i)): 1 \leq i \leq \vert I \vert \}$, where~$t(i)$ denotes the index of the fullest bin that the item~$i$ can be packed satisfying the specified constraints. A relaxed version of this problem consists of finding a mapping between~$I$ and~$\mB$, such that~$t(i) \leq t(j)$ for~$i < j$ and~$i$ cannot fit in~$t(i) + 1$, i.e., ignoring the color constraint and that~$i$ can be packed in~$B' \in \mB$. An algorithm to solve this relaxed problem, with time complexity~$\Theta( \vert I \vert  + \vert \mB \vert)$, is presented in the first part of Algorithm~\ref{alg:MoveItem}.

\begin{algorithm}[b!]     \small
    \SetKwInOut{Input}{input}\SetKwInOut{Output}{output}
    \Input{The sets of items~$I$ and bins~$\mB$, both sorted in non-increasing order}
    \Output{The set of pairs~$(i, t(i))$, for~$1 \leq i \leq \vert I \vert~$}

    \BlankLine
    \tcp{Find a relaxed solution.}

    $to \gets$ vector of size~$ \vert I \vert~$

    $last \gets 0$

    \For{$i \gets 1$ to~$ \vert I \vert~$}{
        \While{$last + 1 \leq \vert \mB \vert$ \textbf{and}~$w_i \leq res(last + 1)$}{
            $last \gets last + 1$ \tcp{This is executed at most~$O(\vert \mB \vert)$ times.}
 }
        $t(i) \gets last$
 }

    \BlankLine
    \BlankLine
    \tcp{Compute the vector~$prev$.}

    $prev \gets$ vector of size~$\vert \mB \vert$ filled with~$0$

    \For{$j \gets 2$ to~$\vert \mB \vert$}{
        \tcp{If~$B_j$ has a tight color and this color is equal to the tight color of~$B_{j-1}$.}
        \eIf{$tight(j) \neq \emptyset$ \textbf{and}~$tight(j) = tight(j - 1)$}{
            $prev[j] \gets prev[j - 1]$ \tcp{ the answers for~$j - 1$ and~$j$ are equal}
 }{
            $prev[j] \gets j - 1$ \tcp{the previous bin is the answer for~$j$.}
 }
 }
    \BlankLine
    \BlankLine
    \tcp{Fix the relaxed solution.}

    \For{$i \gets 1$ to~$ \vert I \vert~$}{
        \For{$k \gets 1$ to~$2$}{
            \If{$t(i) = B(i)$}{
                $t(i) \gets t(i) - 1$
 }
            \If{$t(i) \neq 0$ \textbf{and}~$tight(t(i)) = c_i$}{
                $t(i) \gets prev[t(i)]$
 }
 }
 }
  \caption{\label{alg:MoveItem}Auxiliary Algorithm.}
\end{algorithm}

Subsequently, we outline the procedure for correcting the invalid responses generated by the relaxation in time complexity~$\Theta( \vert I \vert  + \vert \mB \vert)$. Note that if the answer for~$i$ is invalid, we must decrement~$t(i)$ until we find the first valid index, as the relaxation guarantees that~$i$ cannot fit in any bin with an index greater than~$t(i)$.

Invalid answers manifest in two scenarios. Firstly, when~$t(i)$ equals~$B(i)$, we can correct it by decrementing~$t(i)$.  Secondly, if~$tight(t(i)) = c_i$ (and~$i$ is assigned to a bin already tight for its color), then we must verify if another bin can accommodate~$i$. If so, we select the bin with the highest index~$r$. Notably, if this bin exists, its index is the first one from~$t(i) - 1$ down to 1 with a \emph{tight} color different from~$c_i$. If not, we designate that~$t(i) = r = 0$ (an invalid index). To make this process of finding~$r$ efficient, we precompute a vector~$prev$ with all answers using dynamic programming.  Specifically, for each bin index~$s$,~$prev[s]$ is the first preceding bin index (from right to left) where~$tight(s) \neq tight(prev[s])$. The second part of the Algorithm~\ref{alg:MoveItem} delineates the computation of~$prev$ in~$\vert \mB \vert$ steps.

Finally, each invalid response can be corrected at most three times, as the first case can occur only once, and the second case cannot happen consecutively. Thus, we can conclude that the Auxiliary Algorithm shown in the Algorithm~\ref{alg:MoveItem} has time complexity~$\Theta( \vert I \vert  + \vert \mB \vert)$. We say that the set~${\{(i, t(i)): t(i) \neq 0 \wedge 1 \leq i \leq \vert I \vert\}}$ are the moves found by the Auxiliary Algorithm, and this set is used by our following algorithms.

\section{Initial Heuristics\label{section:Initial}}

In this section, we introduce five fast CBPP heuristics. The solutions derived from these heuristics serve as the initial solutions for the VNS and the matheuristic.

\subsection {Best Fit Decreasing}
For BPP, the Best Fit heuristic processes the items in a given permutation and packs the current item in the bin with the lowest residual capacity. A new bin is created if the current item cannot fit into any existing bin. The Best Fit Decreasing (BFD) heuristic operates similarly to the Best Fit, except that the items are processed in a non-increasing order of weight.

A straightforward adaptation of both heuristics for the CBPP is to pack the current item in the fullest bin so that it fits and does not violate the color constraint.

\subsection {Good Ordering}
To illustrate that the BFD has no constant approximation factor for the CBPP, a lower bound instance is to take bin capacity~$W = \frac{3n}{2}$, with~$\frac{n}{2}$ items of weight~$2$ and color~$0$, and~$\frac{n}{2}$ items of weight~$1$ and color~$1$. Despite an optimal solution requiring only~$1$ bin for these instances, the solution generated by the BFD utilizes~$\frac{n}{2}$ bins. Consequently, the non-increasing order may result in a poor performance for Best Fit. To address this issue, we propose the Good Ordering (GO) heuristic, which combines the Best Fit heuristic with a \emph{good ordering} strategy.

We construct the \emph{good ordering}~$R$ incrementally by selecting an item to append to the end of~$R$ in each step. Let~$S$ be the set of non-packed items,~$g$ be one of the most frequent colors of~$S$,~$S_g$ be the subset of~$S$ of items with color~$g$, and~$S_o$ be the subset of~$S$ of items with a color different from~$g$. At each step, we evaluate whether~$ \vert S_g \vert  >  \vert S_o \vert  + 1$. If this condition holds, we select the heaviest item from~$S_g$. Otherwise, we choose the heaviest item with a different color than the last item of~$R$. Note that this algorithm finds the optimal solution for the unfavorable scenario mentioned for BFD\@.

\subsection{Minimum Bin Slack}
The Minimum Bin Slack (MBS) is a BPP heuristic proposed by~\cite{Gupta_1999}. In each step of the MBS, we create a bin~$B'$ with the subset of remaining items that best fill it. To find this subset, we enumerate all valid sets, using the items in a non-increasing order of weight. The MBS' is a proposed modification by~\cite{Fleszar_2002} that replaces the enumeration with Branch-and-Bound and forces the heaviest remaining item to be part of the chosen subset.

We also explore the applicability of MBS' to the CBPP\@. For the adaptation of this algorithm, it would be sufficient to impose that all partial solutions of the Branch-and-Bound respected the color constraint, i.e., each subset should be a valid pattern for CBPP (defined in Section~\ref{section:CBPP}). However, this proved to be slow and led to low-quality solutions.

Given this observation, we have implemented two modifications. The first one imposes a limit of~$n\log n$ iterations to choose each pattern, where an iteration corresponds to an attempt to pack an item into the bin, thus enhancing the execution time. Upon reaching the iteration limit, we choose the best pattern identified thus far. This iteration limit was chosen empirically and its impact on the quality of the solutions obtained seems insignificant. We refer to the algorithm incorporating this first modification as Modified MBS'~(MMBS').

The second modification focuses on improving the quality of solutions, and we do it by replacing the non-increasing order with the \emph{good ordering}. The algorithm with both modifications is called MMBS'-GO\@.

\subsection{Hard BFD}
The Hard BFD is a novel heuristic designed to pack items using the BFD strategy within open bins. However, if an item~$i$ cannot be accommodated using BFD, we perform an alternative routine. This routine selects an item~$j$, among the items with a color different from~$c_i$, which produces the destination bin~$B'$ with the lowest residual capacity when~$i$ and~$j$ are packed together. We break ties by choosing the item~$j$ with the heaviest weight~$w_j$. If there is no valid move, then~$i$ is packed alone into a new bin.

The pseudocode for the Hard BFD heuristic is presented in Algorithm~\ref{alg:HardBFD}. Since all operations within the while loop have a time complexity of~\(O(n)\), this algorithm has an overall time complexity of~\(O(n^2)\). In this pseudocode, we use the Auxiliary Algorithm to find~\(j\) by employing a trick that creates temporary bins by packing item~\(i\) in all of them. This approach enhances performance compared to the straightforward implementation and the binary search-based implementation, which have time complexities of~\(O(n^3)\) and~\(O(n^2 \log n)\), respectively.

\begin{algorithm}[ht]     \small
    \SetKwFunction{AuxiliaryAlgorithm}{AuxiliaryAlgorithm}
    \SetKwInOut{Input}{input}\SetKwInOut{Output}{output}
    \Input{The set~$I$ of items}
    \Output{A feasible solution~$\mB$}

    \BlankLine
    $\mB \gets \emptyset~$ \tcp{Empty Solution}

    \While{$I$ is not empty}{
 Remove the heaviest item~$i$ from~$I$

        \eIf{$i$ can be packed in~$B' \in \mB$ satisfying the capacity and color constraints}{
 pack~$i$ in the bin~$B'$ satisfying these conditions with the lowest residual capacity
 }{
            $\mB' \gets$ temporary bins created by adding~$i$ in each~$B' \in \mB$ with enough free space

            $I' \gets$ items of~$I$ with a color different from~$c_i$

            $moves \gets$ \AuxiliaryAlgorithm{$I'$,~$\mB'$}

            \eIf{$moves \neq \emptyset$}{
 Let~$(j, B'_r)$ be a move in~$moves$ such that~$res(r) - w_j$ is minimum, breaking ties by choosing the heaviest~$w_j$
                
 Let~$B'$ be the bin in~$\mB$ that originated~$B'_r$

 Pack the items~$i$ and~$j$ in~$B'$, and remove~$j$ from~$I$
 }{
 Pack~$i$ in a new bin~$B'$, and add~$B'$ in~$\mB$
 }
 }
 }
  \caption{\label{alg:HardBFD}Hard BFD\@.}
\end{algorithm}

\subsection{Two-by-Two}
Next, we propose a heuristic called Two-by-Two, which operates through an iterative process. In each iteration, we create a bin~$B_r$ and make moves, i.e., choose items to be packed, that minimize a function~$f$. First, we select an item to pack in~$B_r$. Subsequently, at each step, the algorithm evaluates the optimal move between packing one or two items into~$B_r$. If no further valid moves are available, a new iteration commences.

Let~$S$ be the set of non-packed items,~$g$ be one of the most frequent colors of~$S$,~$S_g$ be the subset of~$S$ of items with color~$g$, and~$S_o$ be the subset of~$S$ of items with a color different from~$g$. The function~$f$ is defined as \[f(i, j) = {\left(\frac{res(r) - w_i - w_j}{W}\right)} ^ 2 +  \vert S^{ij} \vert  \cdot {\left( \frac{ \vert S_g^{ij}  \vert }{ \vert S^{ij}  \vert } - \frac { \vert S_g^0 \vert } { \vert S^0 \vert } \right)}^2,\]
where~$S^0$ and~$S_g^0$ are the sets of~$S$ and~$S_g$ at the beginning of the algorithm, and~$S^{ij}$ and~$S_g^{ij}$ are the sets of~$S$ and~$S_g$ after the removal of items~$i$ and~$j$. In the first term of~$f$, the factor~$((res(r) - w_i - w_j) / W)^2$ prioritizes the moves that best fill the bin~$B_r$. The second term of~$f$ penalizes the color discrepancy. Intuitively, in an optimal solution, if we select a random sample of bins, the proportion of items with color~$g$ in this sample should closely resemble~$\vert S_g ^0 \vert /  \vert S ^ 0 \vert$. Thus, the second term penalizes deviations from this proportion. The multiplicative factor~$ \vert S \vert~$ is essential because, without it, the penalty becomes relevant only in the later iterations of the algorithm.

To determine the move that minimizes~$f$ using only one item, we simply enumerate them, a process with a time complexity of~$O(n)$. In this scenario, we evaluate~$f$ considering~$j = \emptyset$ and~$w_j = 0$. Conversely, to identify the two items~$i$ and~$j$ that minimize~$f$, we can enumerate the possibilities, deciding whether~$i \in S_g$ or~$i \in S_o$, doing the same for~$j$. In this way, the second term of~$f$ is constant, and the pair~$(i, j)$ that minimizes~$f$ is the one that best fills the bin. We can use the Auxiliary Algorithm to determine this pair. Algorithm~\ref{alg:TwoByTwo} outlines the pseudocode for this heuristic, with time complexity~$\Theta(n ^ 2)$.

\begin{algorithm}[ht]     \small
    \SetKwFunction{AuxiliaryAlgorithm}{AuxiliaryAlgorithm}
    \SetKwInOut{Input}{input}\SetKwInOut{Output}{output}
    \Input{The set~$I$ of items}
    \Output{A feasible solution~$\mB$}

    \BlankLine
    $\mB \gets \emptyset~~\#$ Empty Solution

    \While{$ \vert I \vert  > 0$}{
        $\bar{B} \gets$ new bin with the item~$i$ that minimize~$f$

 Remove~$i$ from~$I$

        \While{$ \vert I \vert  > 0$}{
            $bestMove \gets~$ the best move with one item (if it exists)

            \ForEach{\normalfont{possibility of}~$S_1, S_2 \in \{S_g, S_o\}$}{

                $\mB' \gets$ temporary bins created by adding independently each item~$i \in S_1$ in~$\bar{B}$

                $moves \gets$ \AuxiliaryAlgorithm{$S_2,$~$\mB'$}

 Evaluate each move found on the two lines above and update~$ bestMove$
 }
            \eIf{$bestMove \neq \emptyset$}{
 Make the best move, remove used items from~$I$, and update~$S, S_g$, and~$S_o$.
 }{
                $\mB \gets \mB \cup \{\bar{B}\}$

                \textbf{break}
 }
 }
 }
  \caption{\label{alg:TwoByTwo}Two-by-Two.}
\end{algorithm}

\section{Variable Neighborhood Search\label{section:VNS}}

The Variable Neighborhood Search~(VNS) is a meta-heuristic proposed by~\cite{Mladenovic_1997} comprises an objective function~$H$, a set of neighborhoods~$\mathcal{N} = \{\mathcal{N}_1, \dots, \mathcal{N}_k\}$, and a \emph{Shake} function. Next, we present these components in more detail.

\subsection{Objective function\label{section:obj}} 
Our objective function~\( H \) operates on two levels, utilizing a \emph{residual vector} composed of the residual capacities of used bins arranged in non-decreasing order. The primary objective is to minimize the number of bins used, while the secondary objective treats the residual vector as a string of integers, prioritizing solutions with the lowest lexicographical order.

\subsection{Neighborhoods}
Let~$I$ denotes the set of all items, and let~$\mB = \{B_1, \ldots, B_k\}$ represents a feasible solution. The neighborhood~$\mathcal{N}_k(\mB)$ encompasses all feasible solutions reachable from~$\mB$ by executing a specific type of move. Each element within this set is referred to as a \emph{neighbor} of~$\mB$.

The VNS conducts local searches in neighborhoods, as elaborated below. One crucial aspect is the heuristic employed in the local search process. In our VNS, we adopt the \emph{Best Improvement} heuristic, which analyzes all neighbors of a solution and transitions to the best-evaluated neighbor based on the objective function~$H$.

Our VNS has neighborhoods with the moves: Move-Item, Swap-Items, Swap-and-Move, and Move-Two-to-One. As the number of items changed in one move is constant for these neighborhoods, the number of elements modified in the residual vector (of the second objective function) is also constant. Since we can compare two neighborhoods based on their objective function variations, this comparison can be done with a constant number of steps. Next, we outline our neighborhoods, elucidating optimizations aimed at achieving efficient algorithms, even for large instances.

\subsubsection{Move-Item Neighborhood}
The Move-Item Neighborhood involves moving an item from one bin to another. An item~$i$ can be moved to bin~$B_r$ if the following conditions are met:~$w_i \leq res(r)$,~$tight(r) \neq c_i$,~$B_r \neq B(i)$, and~$B(i) \setminus \{i\}$ adheres to the color constraint. In such cases, we deem~$(i, B_r)$ a valid move.

For the CBPP, one potential algorithm for this neighborhood entails adapting the approach utilized by~\cite{Fleszar_2002} for the BPP, which exhibits a time complexity~$O(n ^ 2)$. In this algorithm, for each item~$i$, we iterate through the given bins in a non-ascending order of residual capacity. Then, we check if the move~$(i, B_r)$ is valid, where~$B_r$ is the current bin. If so, we evaluate this move. Otherwise, we follow to the next bin.  We can terminate the search upon encountering the first bin~$B_r$ for which~$w_i > res(r)$, as the ordering ensures that the remaining bins lack sufficient space for item~$i$.

However, it is unnecessary to assess all moves to find the best one. Note that, we can identify the best move for each item~$i \in I$ by employing the Auxiliary Algorithm using~$I$ and the current solution~$\mB$ as input. Consequently, the best move for this neighborhood is the best-evaluated one among those returned by the Auxiliary Algorithm. This approach implements the Move-Item Neighborhood with a time complexity of~$\Theta(n)$.

\subsubsection{Swap-Items Neighborhood}
The Swap-Items Neighborhood involves swapping two items from different bins. We can swap two items~$i$ and~$j$, with~$w_i \leq w_j$, between the bins~$B(i)$ and~$B(j)$ if the following conditions are met:~$B(i) \neq B(j)$,~$res(B(i)) + w_i - w_j \geq 0$, and the bins maintain adherence to the color constraint following the swap. In this case, we denote~$(i, j)$ as a valid move.

\cite{Fleszar_2002} propose, for the BPP, a Swap-Items Neighborhood with time complexity~$O(n + M)$, where~$M$ is the number of valid moves, which can be very fast in some solutions with little free space. We present the pseudocode of this neighborhood in the Algorithm~\ref{alg:SwapItemOld}.

\begin{algorithm}[ht]     \small
    \SetKwInOut{Input}{input}\SetKwInOut{Output}{output}

    \Input{The set~$I$ of items sorted by non-decreasing weight}
    \Output{The best move seen}

    \BlankLine
    \For{$i$ \textbf{from}~$1$ \textbf{to}~$ \vert I \vert~$}{

        \For{$j$ \textbf{from}~$i + 1$ \textbf{to}~$ \vert I \vert~$}{

            \If{$res(B(i)) + w_i - w_j < 0$}{

                \textbf{break}
 }

 check if the move~$(i, j)$ is valid, and evaluate it if so
 }
 }
  \caption{\label{alg:SwapItemOld}Swap-Items Algorithm --- Fleszar and Hindi's Version.}
\end{algorithm}

For the CBPP, we could utilize the same algorithm, but it is possible to do something slightly better. To enhance its efficiency, we first sort the items by non-decreasing weight, with items of equal weight arranged by their residual capacities~$res(B(i))$ in non-decreasing order. With this sorted setup, we do something analogous to Algorithm~\ref{alg:SwapItemOld}, but using a better starting point for~$j$. Given an~$i$ fixed, we set~$j$ equal to the largest value~$k$ where~${res(B(i)) + w_i - w_k \geq 0}$, which can be found using a binary search. Subsequently, we iterate decreasing~$j$, checking if the move~$(i, j)$ is valid, and evaluate it. We stop when we find the first move that improves the objective function.

Although the time complexity of this algorithm is~$O (n \log n + M)$,  our tests have demonstrated its superior performance over Fleszar and Hindi's algorithm, particularly in large instances. Algorithm~\ref{alg:SwapItem} delineates its pseudocode, and Lemma~\ref{lemma::swap} attests that the moves considered encompass the best move for our objective function~$H$.

\begin{algorithm}[ht]     \small
    \SetKwInOut{Input}{input}\SetKwInOut{Output}{output}
    \Input{The set~$I$ of items sorted by non-decreasing weight with ties broken by residual capacities~$res(B(i))$ in non-increasing order}
    \Output{The best move among the stored moves}

    \BlankLine
    \For{$i$ \textbf{from}~$1$ \textbf{to}~$ \vert I \vert~$}{

        $j \gets$ greatest~$k$ such that~$res(B(i)) + w_i - w_k \geq 0$

        \While{$w_i \neq w_j$}{

            \If{$(i, j)$ is a valid move that improves the objective function~$H$}{

 store~$(i, j)$

                \textbf{break}
 }

            $j \gets j - 1$

 }
 }
  \caption{\label{alg:SwapItem} Swap-Items Algorithm --- Our Version.}
\end{algorithm}

\begin{lemma}
 Algorithm~\ref{alg:SwapItem} finds the best move for the objective function~$H$ presented in Section~\ref{section:obj}.
    \begin{proof}
 Observe that, for each item~$i$, we only consider swapping it with items~$j$ where~$i < j$. This is because if the best move involves swapping items~$(i, k)$ with~$i > k$, this move will be identified when analyzing item~$k$. Below, we prove that, for a given item~$i$, using the proposed ordering, 
 the largest~$j$ such that the objective function~$H$ improves is the best swap~$(i, j)$ to be made.

 Let~$m = (i, j)$ be a valid swap move,~$B(i)$ be the bin that item~$i$ is packed into before the move~$m$, and~$B(i)^m$ be the bin~$B(i)$ after the move~$m$, i.e.,~$B(i)$ after removing~$i$ and adding~$j$. We can assume~$w_i < w_j$, as moves where~$w_i = w_j$ do not alter the objective function. Since neither~$B(i)^m$ nor~$B(j)^m$ can become empty after the swap, the only way to improve the objective function is through the secondary objective~$h_2$ of~$H$, which seeks to minimize the lexicographical order of the residual vector. 

 For readability, we denote hereafter~$res(B(i))$ and~$res(B(i)^m)$ by~$r_i$ and~$r_i^m$, respectively. After making a move~$m$, we replace~$r_i$ and~$r_j$ with~$r_i^m$ and~$r_j^m$ in the residual vector. Also, note that,~$r_i^m = r_i + w_i - w_j$ and~$r_j^m = r_j + w_j - w_i$. Thus,~$r_i^m < r_i$ and~$r_j^m > r_j$ since~$w_j - w_i > 0$. Also,~$m$ improves~$h_2$ if, and only if,~$\min(r_i^m, r_j^m) < \min(r_i, r_j)$, which happens if, and only if,~$r_i^m < \min(r_i, r_j)$.

 Suppose that there is another swap move~$o = (i, k)$ that improves the objective function. Thus, we have that~$r_i^o < \min(r_i, r_k)$ and~$r_k^o > r_k$. 
    
 In order to compare~$m$ and~$o$, we must consider only the triple of elements~$\{r_i, r_j, r_k\}$ of the residual vector (possibly~$B(j) = B(k)$, but the argument is analogous). The move~$m$ replaces this triple with~$M = \{r_i^m, r_j^m, r_k\}$ and the move~$o$ replaces this triple with~$O = \{r_i^o, r_j, r_k^o\}$. 
    
 First, suppose that~$w_j < w_k$. As~$r_i^o = r_i + w_i - w_k$, we have that~$r_i^o < r_i^m$. Since~$r_i^o < r_k < r_k^o$,~$r_i^m < r_j < r_j^m$ and~$r_i^o < r_i^m$, it follows that~$r_i^o$ is strictly smaller than all other elements of both triples~$M$ and~$O$. Therefore, the move~$o$ is better than~$m$ to minimizing~$h_2$. 
    
 Finally, suppose that~$w_j = w_k$ and~$r_j < r_k$ (if~$r_j = r_k$, then both moves are equal under~$h_2$). Thus,~$r_i^m = r_i^o$ and~$r_j^m = r_j + w_j - w_i < r_k + w_k - w_i = r_k^o$. As~$r_j < r_k < r_k^o$ and~$r_j < r_j^m$, move~$o$ is better than~$m$ to minimize~$h_2$.
    
 Therefore, if~$m = (i, j)$ and~$o = (i, k)$ are swaps that improve the objective function and~$j < k$, then~$o$ is better than~$m$. Thus, the largest~$j$ such that the objective function~$H$ improves is the best swap~$(i, j)$ to be made. 
    
 Consequently, the best move for the whole neighborhood is determined by selecting the best-evaluated move among these moves. Thus, Algorithm~\ref{alg:SwapItem} is correct.
    \end{proof}
    \label{lemma::swap}
\end{lemma}

\subsubsection{Move-Two-to-One Neighborhood}
The Move-Two-to-One Neighborhood involves selecting two items from different bins to move to a third bin. The straightforward implementation of this neighborhood has time complexity~$O(n ^ 3)$. However, there is a more efficient approach.  We can select an item~$i$ and move it to all other bins~$\mB \setminus B(i)$, creating a set~$\mB'$ of temporary bins. Afterward, we utilize the Auxiliary Algorithm on sets~$I \setminus \{i\}$ and~$\mB'$, determining the best destination bin for each pair of items~$(i', j)$, where~$i' = i$ and~$j \in I \setminus \{i\}$. By repeating this process for each item~$i \in I$, we have an algorithm for this neighborhood with a time complexity of~$\Theta(n^2)$.

\subsubsection{Swap-and-Move Neighborhood}
The Swap-and-Move Neighborhood involves selecting three items~$i, j, k$ from different bins, swapping~$i$ and~$j$, and moving~$k$ to~$B'(i)$, where~$B'(i)$ is the bin~$B(i)$ after removing~$i$ and adding~$j$. This neighborhood has time complexity~$O(n^3)$ when implemented in the completed way, which offers a low cost-benefit. Thus, we adopt a constrained version.

Instead of considering all possible candidates for~$k$, we focus solely on the heaviest item that can fit into~$B'(i)$ while also satisfying the color constraint. Let~$I'$ be the subset of~$I$ that can be removed from their bins without violating the color constraint, sorted in a non-decreasing order of weight. We then use a binary search in~$I'$ to find the greatest index~$l$ such that~$w_{I'[l]} \leq res(B'(i))$. Note that~$I'[l]$ serves as the initial candidate for~$k$, but it may not meet the color constraint or be packed in the same bin as~$i$ or~$j$.

Similar to the Auxiliary Algorithm, we address the potential occurrence of an invalid solution using the same strategy with the vector~$prev$. The Auxiliary Algorithm's task is to find a bin for each item, indexing the vector~$prev$ by bins. However, in this case, we need to find an item~$k$ to pack in each bin, so we must index the vector~$prev$ by the items~$I'$. Specifically, each position~$prev[l]$ stores the nearest previous position~$m$ where the tight color of bin~$B(I'[l])$ differs from that of bin~$B(I'[m])$, i.e.,~$tight(B(I'[l])) \neq tight(B(I'[m]))$.

Thus, for each valid swap between~$i$ and~$j$, we find the index~$l$ and check if~$I'[l]$ can be packed in~$B'(i)$. If so, we found~$k$. Otherwise, either~$I'[l]$ belongs to the same bin as~$i$ or~$j$, in which case we decrement~$l$; or~${tight(B'(i)) = tight(B(l))}$, and in this case, we set~$l = prev[l]$. Unfortunately, the first case now can occur~$O(\vert B(i) \cup B(i )\vert)$ times, as all items of~$B(i)$ and~$B(j)$ can belong to~$I'$.

Since finding the item~$k$ of each swap takes~$ O (\log \vert I' \vert +  \vert B _{\max} \vert)$, where~$\vert B _{\max} \vert~$ represents the maximum number of items in a bin, and there are~$ O(n^2)$ swaps, we have an algorithm with time complexity~${O(n ^2 \cdot (\log n +  \vert B_{\max} \vert))}$. Algorithm~\ref{alg:SwapAndMove} outlines the pseudocode of this neighborhood.

\begin{algorithm}[ht]     \small
    \SetKwInOut{Input}{input}\SetKwInOut{Output}{output}
    \Input{The set~$I$ of items sorted by non-decreasing weight and a solution~$\mB$}
    \Output{The best move among the stored moves}

    \BlankLine

    $I' \gets$ the items of~$I$ that can be removed from their bins without violating the color constraint sorted in a non-decreasing order of weight

    $prev \gets$ array of size~$ \vert I' \vert~$ with each index initialized to~$0$

    \BlankLine
    \For{$i \gets 2$ to~$ \vert I' \vert~$}{

        \eIf{$tight(B(I'[i])) \neq \emptyset$ \textbf{and}~$tight(B(I'[i])) = tight(B(I'[i - 1]))$}{
            $prev[i] \gets prev[i - 1]$
 }{
            $prev[i] \gets i - 1$
 }
 }

    \BlankLine
    \ForEach{swap~$(i, j)$ that satisfies the capacity and color constraints}{

        $l \gets$ greatest~$k$ in~$I'$ such that~$res(B(i)) + w_i - w_j - w_{I'[k]} \geq 0$

        $t_i \gets$ the tight color of~$B(i)$ when replacing the item~$i$ by~$j$

        \While {$l \geq 1$}{
            \If{$B(I'[l]) = B(i)$ \textbf{or}~$B(I'[l]) = B(j)$}{

                $l \gets l - 1$
 }

            \If{$l \geq 1$ \textbf{and}~$t_i = c_{I'[l]}$}{

                $l \gets prev[l]$
 }

            \If{$l \geq 1$ \textbf{and}~$(i, j, I'[l])$ is a valid move}{

 store~$(i, j, I'[l])$

                \textbf{break}
 }
 }
 }
  \caption{\label{alg:SwapAndMove}Swap-and-Move Algorithm.}
\end{algorithm}

\subsection{Shake Algorithm}
The Shake Algorithm is a function that makes random moves, seeking to escape from local optima. In particular, our Shake Algorithm chooses to execute one of two subroutines with equal probability. 

In the first subroutine, we have a candidate list~$L$ of items, where initially~$L = I$. Thus, we randomly select an item~$i \in L$ and build a list of valid moves~$M$ considering moving~$i$ to all other bins and swapping~$i$ with all other items~$j$ such that~$B(i) \neq B(j)$. If~$M$ is not empty, we uniformly at random choose a move of~$M$ and perform it, removing the items involved from~$L$. Otherwise, we only remove~$i$ from~$L$. This process is repeated until we make 20 successful moves or~$L$ is empty.

In the second subroutine, we uniformly at random choose two bins and repack their items using the Best Fit heuristic in a random order.

It is important to emphasize the aggressive nature of our shake algorithm, as its moves have the potential to significantly alter the solution to a much worse solution. Our algorithm is based on the one proposed by~\cite{Fleszar_2002} for the BPP, which is equal to the first subroutine without the constraint that bins must be distinct in different moves. The second routine was added since the perturbation generated by the first one can be small in instances with little free space.

\subsection{The Algorithm}
Given an objective function~$H$, a set of neighborhoods~$\mathcal{N} = \{\mathcal{N}_1, \dots, \mathcal{N}_k\}$, and a \emph{Shake} function, our VNS Algorithm is described as follows. We start with~$k = 1$ and a feasible solution~$\mB$ constructed using some initial heuristics outlined earlier. Thus, at each step, we conduct a local search in~$\mB$ with the neighborhood~$\mathcal{N}_k$ until we arrive at a solution~$\mB'$, which is locally optimal. If~$\mB'$ is better than~$\mB$, then we make~$\mB\gets \mB'$ and~$k \gets 1$. Otherwise, we increment~$k$ by 1. If we reach~$k = \vert \mathcal{N} \vert + 1$, indicating that we have traversed all the neighborhoods in~$\mathcal{N}$, we run the Shake Algorithm to generate a random perturbation in~$\mB~$, and we start a new iteration with~$k = 1$.

To prevent indefinite execution, we implement two stopping criteria. The first criterion is if the solution reaches the lower bound~$L_1$, where~${L_1 = \lceil\sum_{i \in I}w_i / W \rceil}$, which is the minimum number of bins to pack the items if they were fragmentable. The second criterion is an execution time limit~$t_{\max}$. Upon meeting any of these criteria, the algorithm returns the best between the initial solution and the solutions at the end of each local search. The pseudocode of our VNS is provided in Algorithm~\ref{alg:VNS}. Note that the internal \emph{while} is the Variable Neighborhood Decreasing~(VND) meta-heuristic, which deviates from the conventional VNS approach that employs the Shake Algorithm before each local search. This VNS version is also used by~\cite{Fleszar_2002} and yielded superior results in our experiments.

\begin{algorithm}[ht]     \small
    \SetKwFunction{ShakeAlgorithm}{ShakeAlgorithm}
    \SetKwFunction{BestImprovement}{BestImprovement}
    \SetKwInOut{Input}{input}\SetKwInOut{Output}{output}
    \Input{The set of neighborhoods~$\mathcal{N}$, the objective function~$H$, and a feasible solution~$\mB$}
    \Output{The incumbent solution}

    \BlankLine
    $incumbent \gets \mB$

    \While{do not reach any stopping criteria}{

        $k \gets 1$

        \While{do not reach any stopping criteria \textbf{and}~$k \leq \vert \mathcal{N} \vert~$}{

            $\mB' \gets$ \BestImprovement{$\mathcal{N}_k$,~$\mB$}

            \eIf{$H(x') < H(x)$}{

                $\mB \gets \mB'$

                $k \gets 1$

                \If{$H(x) < H(incumbent)$}{

                    $incumbent \gets \mB$
 }
 }{
                $k \gets k + 1$
 }
 }

        $\mB \gets$ \ShakeAlgorithm{$\mB$}
 }
  \caption{The VNS Algorithm\label{alg:VNS}.}
\end{algorithm}

\subsection{Neighborhood Ordering}
Note that the VNS performs a local search in the neighborhood~$\mathcal{N}_k$ only if the current solution is a local optimum in the neighborhoods~$\{\mathcal{N}_1, \dots,  \mathcal{N}_{k - 1}\}$. Therefore, to save computational time, low-cost neighborhoods usually precede high-cost neighborhoods. In our VNS, the best results were using the neighborhoods in the same order in which they were presented.

\section{Matheuristic\label{section:MH}}
Now, we present a matheuristic (MH) that combines linear programming with the meta-heuristics VNS and GRASP, where GRASP is a semi-greedy meta-heuristic initially described by~\cite{Resende_1989}. 

Our MH operates through a procedure divided into three stages: A, B, and C, which are executed alternately until a solution with a value equal to the lower bound~$L_1$ is found or until the time limit~$t_{\max}$ is reached. Throughout these stages, we maintain an incumbent solution~$F_{best}$, representing the best solution known, and a pattern set~$\mathcal{P}'$. The set~$\mathcal{P}'$ consists of patterns from some generated solutions, i.e., the algorithm selects a set of solutions~${\mathcal{S} = \{S_1, \ldots S_h\}}$, and if a pattern~$p$ appears in any solution~$S_i \in \mathcal{S}$, then~$p \in \mathcal{P}'$. In the first iteration,~$F_{best}$ comes from the Two-by-Two heuristic (because of its performance, as presented in Section~\ref{section:Results}) and~$\mathcal{P}' = F_{best}$. Next, we explain the three stages in detail.

In Stage A, we run VNS for~$t_A$ seconds, using the solution~$F_{best}$ as the initial solution. Additionally, whenever we reach a local optimum~$S$ within a neighborhood, we add the bins of~$S$ to~$\mathcal{P}'$. The objective of this stage is to collect good patterns.

In Stage B, we solve the restricted linear relaxation of the formulation presented in Section~\ref{section:CBPP}. This involves solving the linear program obtained by removing the integrality constraint on~$\lambda$ and considering only the variables associated with the subset of patterns~$\mathcal{P}' \subseteq \mathcal{P}$. The goal at this stage is to filter the patterns in~$\mathcal{P}'$, based on the intuitive notion that patterns used, either fully or partially, by the optimal solution~$\lambda^*$ of this restricted model should form a high-quality integer solution.

In Stage C, we execute GRASP for~$t_C$ seconds, aiming to round the relaxation by combining greedy and random aspects. GRASP involves a procedure with multiple restarts, comprising two steps: building a solution and performing a local search.

In the building step, we start with the optimal solution~$\lambda^*$ of the linear relaxation and a parameter~$0 < \alpha < 1$. Let~$S^*$ be the set of patterns~$p$ with~$\lambda^*_p > 0$. We incrementally build a solution~\( F \), initially setting~\( F = \emptyset \). In each iteration, we create a candidate list~$L \subseteq S^*$, where~$ \vert L \vert  = \lceil \alpha \vert S^* \vert  \rceil$ and~$L$ is composed of the patterns~$p$ with the largest values of~$\lambda_p^*$. We iterate over the patterns of a random ordering of~$L$ and add the current pattern~$p$ to~$F$ if~$p \cap p' = \emptyset$ for each pattern~$p' \in F$. After processing all patterns of~$L$, we remove them from~$S^*$ and start a new iteration.

At the end of this process, a subset~$I' \subset I$ of items might remain unpacked in any pattern in~$F$, even after analyzing all patterns of~$S^*$. In this case, we pack the items of~$I'$ into~$F$ using a randomized version of the Two-by-Two heuristic. To introduce randomness in the Two-by-Two heuristic, we select a random move from within the top~$\alpha$ proportion of the best-evaluated moves instead of always choosing the best move.

Finally,~$F$ becomes a feasible solution to the problem, and we begin the local search step. We conduct a local search on~$F$ using the VND scheme from our VNS\@. Thus, we store the patterns of~$F$ in~$\mathcal{P}'$ and update~$F_{best}$ if necessary. If the time limit~$t_C$ is reached, we proceed to Stage A; otherwise, we return to the building step.

As an optimization, we have found it advantageous to run the VND only on the part of the solution created by the randomized Two-by-Two heuristic. There are two main reasons for this. First, the patterns of~$S^*$ are usually well-filled bins, making it unlikely that the local search will improve the solution without a perturbation routine like the Shake Algorithm. Second, the local search becomes much faster since the set~$I'$ is typically small compared to~$I$, allowing us to run more iterations of the MH\@.

\section{Computational Results\label{section:Results}}

Below, we provide the instances utilized in our computational experiments\footnote{All instances and source codes are accessible at \url{https://gitlab.com/renanfernandofranco/fast-neighborhood-search-heuristics-for-the-colorful-bin-packing-problem}.}, and we present the results in three stages. Initially, we compare our initial heuristics using performance profiles and statistical tests. In the second stage, we compare the average gap and runtime of both VNS and MH algorithms. Finally, we compare our MH algorithm with the exact algorithms proposed by~\cite{Borges_2023}.

All results were obtained using a computer with processor Intel\textsuperscript{\textregistered}\ Xeon\textsuperscript{\textregistered}\ CPU E5--2630 v4 @ 2.20GHz with 64 GB de RAM, operational system Ubuntu 18.04.6 LTS (64 bits). We used the language C++17 compiled with GCC 7.5.0. The commercial solver Gurobi 10.0.3 was used in \textit{single thread} to solve the linear relaxation. Finally, all our experiments used 0 as the seed for the random number generator.

\subsection{Instances}
The instances used in our experiments are divided into four classes.

The first class is an adaptation of Augmented IRUP~(AI) instances of the BPP to the CBPP, characterized by pairs~$(n, W)$ belonging to~$\{ (202, 2500), (403, 10000), (601, 20000) \}$, where~$n$ is the number of items and~$W$ is the bin capacity. In the context of the BPP, an IRUP instance maintains the Integer Rounding Up Property (IRUP) for the Kantorovich-Gilmore-Gomory Formulation, ensuring that the optimal solution value equals the relaxation value rounded up.

The second class involves an adaptation of FalkenauerT instances from the BPP to the CBPP\@. These instances have~$n \in \{60, 120, 249, 501\}$,~$W = 1000$, and there is always an optimal solution that comprises~$\frac{n}{3}$ bins, each containing three items.

The last two classes consist of randomly generated instances, with~${n \in \{ 102, 501, 2001, 10002\}}$ and~$W \in \{ 10^3 + 1, 10^6+1 \}$.

The first random class, Random10to80 instances, have item weights randomly chosen from the discrete ranges~$(0, \frac{W}{4}]$ and~$[\frac{W}{10}, \frac{4W}{5}]$, respectively.

The second random class, Random25to50 instances, consists of~$\frac{n}{3}$ triples of items with a total weight equal to~$W$, where each item has a randomly chosen weight in the discrete range~$[\frac{W}{4}, \frac{W}{2}]$.

The BPP instance classes AI and FalkenauerT are available in the \hyperlink{http://or.dei.unibo.it/library/bpplib}{BPP Lib} created by~\cite{Delorme_2018}, and their optimal solutions are known. In the AI class, finding an optimal solution is very difficult, but finding a solution with one additional bin is relatively easy. While the FalkenauerT and Random25to50 classes are similar, we introduced the latter due to the absence of instances with a large number of items or bin capacities in FalkenauerT\@.

The items are colored in two different ways. The first type is \emph{Uniform Coloring}, where each instance has a number of distinct colors~$Q \in \{ 2, 5, 20 \}$. The color of each item is then chosen randomly from the discrete range~$[0, Q)$. This coloring method is applied to all instance classes. We take special care with the AI, FalkenauerT, and Random25to50 classes by obtaining an optimal BPP solution and coloring the instance so that this solution remains feasible for CBPP\@. It can be done during the coloring process by preventing an item from being assigned to a color~$c$ if the tight color of its bin is~$c$.

The second type of coloring is called \emph{Q2H}, which is applied to a BPP instance using one of its optimal solutions. This coloring is applied for all classes except the Random10to80 instances, for which optimal BPP solutions are unknown (optimal BPP solutions for other classes are known from the literature or by construction). Given an optimal BPP solution~$\mathcal{B}$, for each bin~$B \in \mathcal{B}$, we denote the heavy items as the heaviest~$\lceil \vert B \vert / 2 \rceil$ items in~$B$, and the light items as the remaining ones. Thus, in this coloring, we color the heavy items with color 1 and the light ones with color 0. The purpose of this coloring is to challenge the initial heuristics. Suppose these heuristics focus on packing the heavy items first or using many light items to fill a bin. In that case, the resulting solution will require significantly more bins than an optimal solution.

We generate CBPP instances for each possible parameter combination of BPP instances. For the Random instances, we generate five instances for each parameter combination.

\subsection{Comparison between Initial Heuristics}
 Next, we compare our initial heuristics using a performance profile and statistical tests.

    \subsubsection{Performance Profile}
 We use performance profiles to compare the initial heuristics, as described below. First, for each instance, we determine how much worse an algorithm's solution is compared to the best solution among all algorithms (referred to as the \emph{reference}). This comparison allows us to generate a graph for each algorithm~$A$, where a point~$(p, \theta)$ indicates the proportion of instances in which~$A$ has solutions with a value at most~$\theta~$ times worse than the reference is equal to~$p$. For instance, if the point~$(\frac{1}{2}, 3)$ appears on the graph for the BFD algorithm, it means that in half of the instances, BFD finds a solution that is at most three times worse than the reference.

\begin{figure}[ht]
\centering
\includegraphics[width=\textwidth]{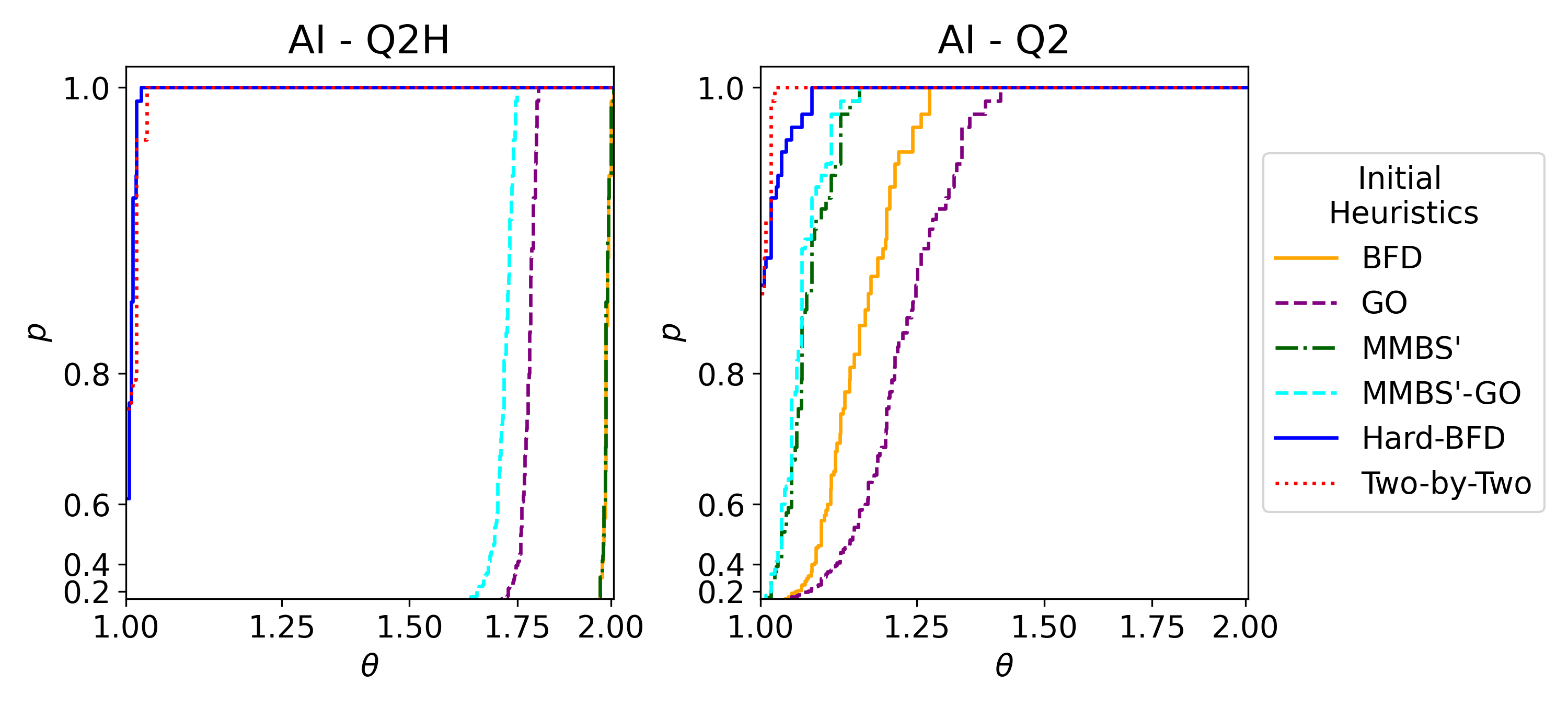}
\caption{Comparison between Initial Heuristics in AI instances.\label{fig:AIInitial}}
\end{figure}

 Figure~\ref{fig:AIInitial} presents the performance profile for the AI class involving Q2 and Q2H coloring. As shown, the heuristic Two-by-Two has the best performance, being twice as good as BFD in some instances with Q2H coloring. The graph for the other coloring is similar to that of Q2, with the difference decreasing as the number of colors increases. This trend is consistent across other instance classes, and we omit the respective graphs.

\begin{figure}[b!]
    \centering
    \includegraphics[width=\textwidth]{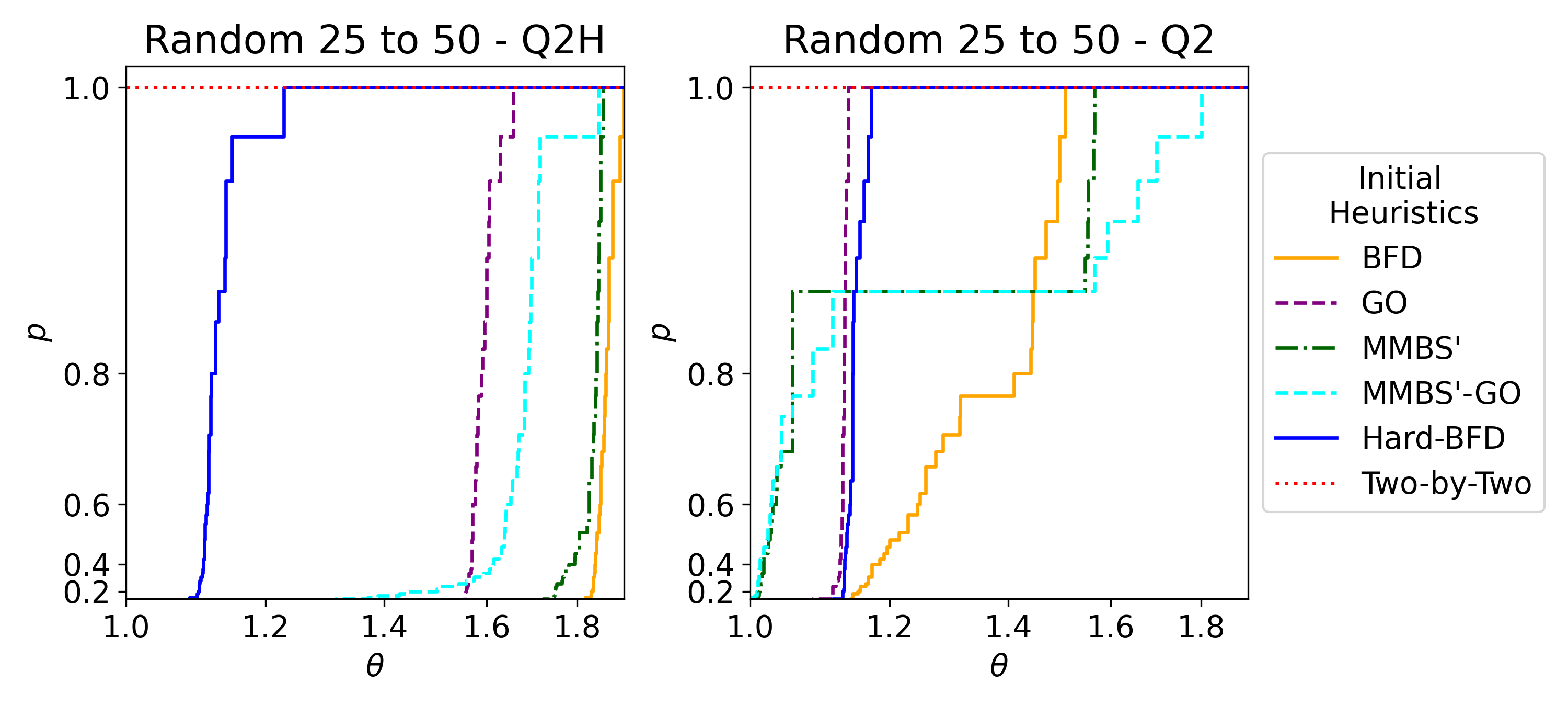}
    \caption{Comparison between Initial Heuristics in Random25to50 instances.\label{fig:25to50Initial}}
\end{figure}

Figure~\ref{fig:25to50Initial} presents the results for Random25to50 instances with Q2 and Q2H coloring. Once again, the Two-by-Two heuristic proves to be the most effective, showing a notable advantage over Hard-BFD, which requires at least 10\% more bins. Interestingly, both versions of MMBS' perform better than Hard-BFD in~90\% of the instances with Q2 coloring. The results for FalkenauerT instances are similar to those for Random25to50, further highlighting the effectiveness of the Two-by-Two heuristic.

\begin{figure}[ht]
\centering
\includegraphics[width=\textwidth]{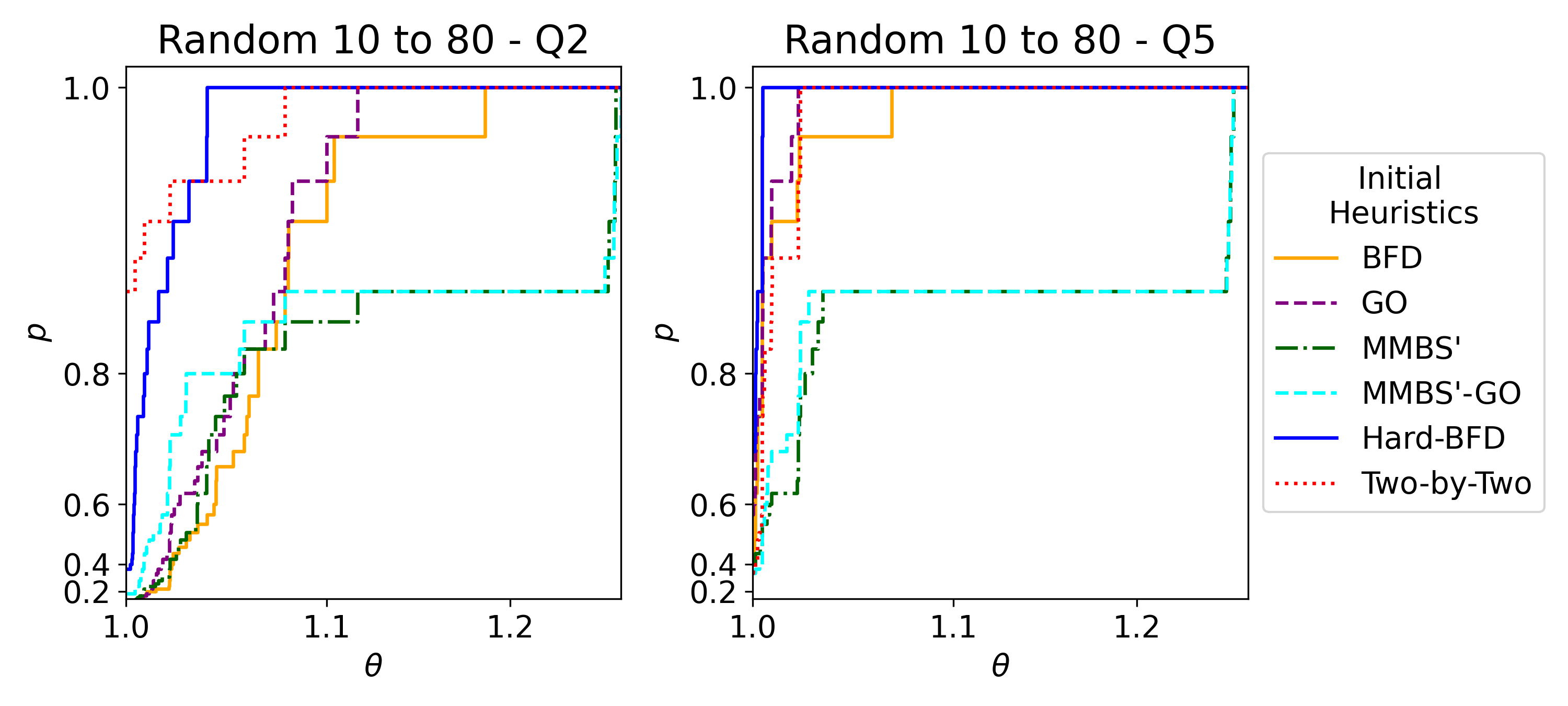}
\caption{Comparison between Initial Heuristics in Random10to80 instances.\label{fig:10to80Initial}}

\end{figure}

Next, Figure~\ref{fig:10to80Initial} shows the performance profile for Random10to80 instances with Q2 and Q5 coloring. With Q2 coloring, there is no clear dominance between the Two-by-Two and Hard-BFD heuristics. Nonetheless, the Hard-BFD heuristic outperforms all others with Q5 coloring, although the difference for the Two-by-Two is not so significant.

Finally, while the Good Ordering heuristic often enhances solution quality, there are exceptions. For instance, the Good Ordering heuristic performs worse than the BFD heuristic in the AI class with Q2, Q5, and Q20 coloring.

\subsubsection{Statistic Tests}
In this section, we conduct a hypothesis test to decide between the null hypothesis~$H_0$ and the alternative hypothesis~$H_1$. The null hypothesis~$H_0$ asserts that there are no significant differences between our initial heuristics. Conversely, $H_1$ suggests that there is a significant difference between at least one pair of heuristics. We use a significance level of~$\alpha = 0.05$, which determines the threshold for rejecting~$H_0$.

Our comparison is based on non-parametric analysis, following the approach of~\cite{Demvsar_2006}. Given that we are comparing multiple strategies across different instances, we employ the adaptation of the Friedman test~\citep{Friedman_1937} as modified by~\cite{Iman_1980}. The null hypothesis is rejected when the test value exceeds the critical value according to the F-distribution. In such cases, we perform a post-hoc analysis using the Nemenyi test~\citep{Nemenyi_1963} to identify groups of statistically equivalent strategies. This is determined by examining whether the average ranks of two samples differ by more than a specified critical difference (CD).

Our test involves~\(k = 6\) algorithms and~\(N = 1200\) instances. For these parameters, the critical value of the F-distribution is 2.22. Let~\(r_i^j\) be the rank of the~\(j\)-th algorithm in the~\(i\)-th instance, where the rank is~\(h\) if the algorithm gives the~\(h\)-th best result (in case of ties, the average is used). Thus, let~\(R_j\) be the average rank for the~\(j\)-th algorithm, defined as~\(R_j = \sum_{i = 1}^N r_i^j / N\) for~\(j \in \{1, \ldots, k\}\). These average ranks~\(R_j\) are used to compute the test value~\(F_F\) for the Iman-Davenport test, where our data yields~\(F_F = 484.02\). Since~\(F_F\) is greater than the critical value, we reject the null hypothesis~\(H_0\) and accept the alternative hypothesis~\(H_1\). Consequently, we can apply the Nemenyi test to determine which heuristics have statistically significant differences.

\begin{figure}[ht]
\centering
\includegraphics[width=0.7\textwidth]{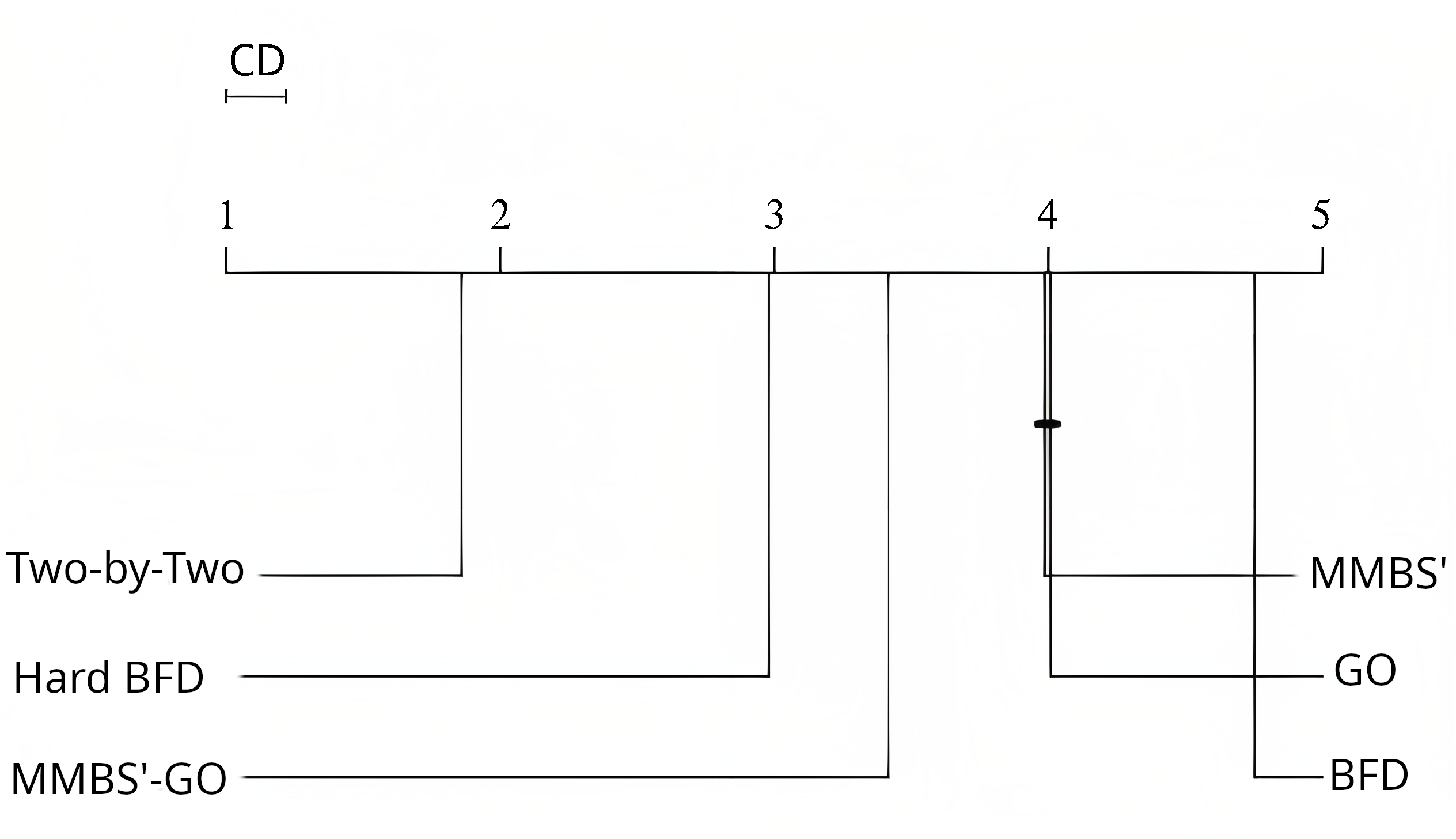}
\caption{Nemenyi Test for the Initial Heuristics.\label{fig:Statistic1}}

\end{figure}

Figure~\ref{fig:Statistic1} presents a graph illustrating the Nemenyi test results. For our data, the CD is~$0.22$. The heuristic average ranks $R_j$ are displayed from left to right, with the best heuristic positioned furthest to the left. If two heuristics have average ranks with a difference less than the CD, they are connected by a horizontal line, indicating no significant statistical difference between them. As shown in the figure, the Two-by-Two heuristic performs the best. Additionally, the \emph{good ordering} technique generally improves performance, as evidenced by MMBS'-GO and GO heuristics outperforming their original versions, MMBS' and BFD\@. Finally, there is no significant statistical difference between the MMBS' and GO heuristics.

\subsection{Comparison between VNS and MH}

Next, we compare the VNS and the MH using a time limit~$t_{\max}$ of 60 seconds. For the MH, we set~$\alpha = 0.3$,~$t_A = 5$ seconds and~$t_C = 1$ second. Both VNS and MH use the Two-by-Two heuristic to generate the initial solution, as it produces the best results. To avoid repetition, we group certain types of coloring and the two types of bin capacity. Moreover, in the AI, FalkenauerT~(FK), and Random25to50 classes, we use an additional coloring called \emph{QN}, where each item has a different color, making it a BPP instance as the color constraint is irrelevant. Since the other coloring in these classes have the same optimal solution value by construction, this \emph{QN} coloring allows us to verify whether the other colorings provide harder or easier instances compared to the BPP version.

\begin{table}[!b]
\resizebox{\columnwidth}{!}{%
\begin{tabular}{ccc|cc|ccc|ccc}
\toprule
                                                    &                                                                                    &                                     &                                     &                                                                         & \multicolumn{3}{|c}{\textbf{VNS}}                                                                                                                                                                                                                              & \multicolumn{3}{|c}{\textbf{MH}}                                                                                                                                                                                                                               \\ \midrule
\textbf{Class}                                                                            & \textbf{N} & \textbf{Coloring} & \textbf{Total} & $\mathbf{\Delta_{Avg}^{Ini}}$ & $\mathbf{\Delta_{Avg}^{Fin}}$ & $\mathbf{\Delta_{Max}^{Fin}}$ & \textbf{t (s)} & $\mathbf{\Delta_{Avg}^{Fin}}$ & $\mathbf{\Delta_{Max}^{Fin}}$ & \textbf{t (s)} \\
\midrule
\multirow{12}{*}{AI}          & \multirow{4}{*}{202} & Q2       & 50    & 1.14   & 1.00          & 1         & 60.00         & \textbf{0.34} & 1         & \textbf{30.43} \\
                              &                      & Q2H      & 50    & 1.78   & 1.00          & 1         & 60.00         & \textbf{0.10} & 1         & \textbf{13.76} \\
                              &                      & Q5 / Q20 & 100   & 1.00   & 1.00          & 1         & 60.00         & \textbf{0.13} & 1         & \textbf{16.16} \\
                              &                      & QN       & 50    & 1.00   & 1.00          & 1         & 60.00         & \textbf{0.10} & 1         & \textbf{17.92} \\ \cline{2-11}
                              & \multirow{4}{*}{403} & Q2       & 50    & 1.16   & 1.00          & 1         & 60.00         & \textbf{0.48} & 1         & \textbf{39.65} \\
                              &                      & Q2H      & 50    & 1.84   & 1.00          & 1         & 60.00         & \textbf{0.38} & 1         & \textbf{29.77} \\
                              &                      & Q5 / Q20 & 100   & 1.00   & 1.00          & 1         & 60.00         & \textbf{0.52} & 1         & \textbf{39.06} \\
                              &                      & QN       & 50    & 1.00   & 1.00          & 1         & 60.00         & \textbf{0.52} & 1         & \textbf{38.34} \\ \cline{2-11}
                              & \multirow{4}{*}{601} & Q2       & 50    & 1.22   & 1.02          & 2         & 60.00         & \textbf{0.94} & 1         & \textbf{58.85} \\
                              &                      & Q2H      & 50    & 2.32   & 1.00          & 1         & 60.00         & \textbf{0.80} & 1         & \textbf{52.63} \\
                              &                      & Q5 / Q20 & 100   & 1.00   & 1.00          & 1         & 60.00         & \textbf{0.95} & 1         & \textbf{58.39} \\
                              &                      & QN       & 50    & 1.00   & 1.00          & 1         & 60.00         & \textbf{0.98} & 1         & \textbf{59.07} \\ \midrule
\multirow{16}{*}{FK} & \multirow{4}{*}{60}  & Q2       & 20    & 1.00   & \textbf{0.00} & 0         & \textbf{0.04} & \textbf{0.00} & 0         & 0.05           \\
                              &                      & Q2H      & 20    & 0.95   & \textbf{0.00} & 0         & \textbf{0.02} & \textbf{0.00} & 0         & \textbf{0.02}  \\
                              &                      & Q5 / Q20 & 40    & 1.00   & \textbf{0.00} & 0         & \textbf{0.06} & \textbf{0.00} & 0         & 0.09           \\
                              &                      & QN       & 20    & 1.05   & \textbf{0.00} & 0         & \textbf{0.06} & \textbf{0.00} & 0         & 0.07           \\ \cline{2-11}
                              & \multirow{4}{*}{120} & Q2       & 20    & 1.00   & \textbf{0.00} & 0         & \textbf{0.16} & \textbf{0.00} & 0         & 0.23           \\
                              &                      & Q2H      & 20    & 1.35   & \textbf{0.00} & 0         & \textbf{0.77} & \textbf{0.00} & 0         & 1.36           \\
                              &                      & Q5 / Q20 & 40    & 1.00   & \textbf{0.00} & 0         & \textbf{0.12} & \textbf{0.00} & 0         & 0.14           \\
                              &                      & QN       & 20    & 1.95   & \textbf{0.00} & 0         & \textbf{0.10} & \textbf{0.00} & 0         & 0.12           \\ \cline{2-11}
                              & \multirow{4}{*}{249} & Q2       & 20    & 1.50   & \textbf{0.00} & 0         & \textbf{0.08} & \textbf{0.00} & 0         & 0.10           \\
                              &                      & Q2H      & 20    & 2.25   & \textbf{0.00} & 0         & \textbf{0.26} & \textbf{0.00} & 0         & 0.35           \\
                              &                      & Q5 / Q20 & 40    & 1.00   & \textbf{0.00} & 0         & \textbf{0.09} & \textbf{0.00} & 0         & \textbf{0.09}  \\
                              &                      & QN       & 20    & 4.05   & \textbf{0.00} & 0         & \textbf{0.08} & \textbf{0.00} & 0         & \textbf{0.08}  \\ \cline{2-11}
                              & \multirow{4}{*}{501} & Q2       & 20    & 1.80   & \textbf{0.00} & 0         & \textbf{0.10} & \textbf{0.00} & 0         & 0.14           \\
                              &                      & Q2H      & 20    & 3.15   & \textbf{0.00} & 0         & 0.47          & \textbf{0.00} & 0         & \textbf{0.41}  \\
                              &                      & Q5 / Q20 & 40    & 1.02   & \textbf{0.00} & 0         & \textbf{0.13} & \textbf{0.00} & 0         & 0.18           \\
                              &                      & QN       & 20    & 8.20   & \textbf{0.00} & 0         & \textbf{0.12} & \textbf{0.00} & 0         & 0.15           \\ \bottomrule 

\end{tabular}
}
\caption{Results for the BPP Instances.}
\label{tab::BPP}
\end{table}

In Tables~\ref{tab::BPP} and~\ref{tab::random}, the instance classes are divided by the number of items~(N) and coloring, and the column \textit{Total} indicates the number of instances in the class. The columns~$\Delta_{Avg}^{Ini}$ and~$\Delta_{Avg}^{Fin}$ indicate the average differences between the lower bound~$L_1$ and the initial and final solution values, respectively. The column~$\Delta_{Max}^{Fin}$ indicates the maximum difference between a final solution and the lower bound~$L_1$. Additionally, the column \textit{t} provides the average time in seconds. The best values for~$\Delta_{Avg}^{Fin}$ and \textit{t} in each row are highlighted in bold.

Our approaches have demonstrated outstanding performance. They find optimal solutions in all Random25to50 instances with 102 and 501 items and achieve solutions with at most one additional bin compared to optimal solutions for 2001 items. Particularly noteworthy is their performance in the class with 10002 items and Q2H coloring, where the average initial difference of 82.1 bins is reduced to at most 1 bin.
\begin{table}[bt]
\resizebox{\columnwidth}{!}{%
\begin{tabular}{ccc|cc|ccc|ccc}
\toprule
                                                                                                    &                                                                                    &                                     &                                     &                                                                         & \multicolumn{3}{c|}{\textbf{\textbf{\textbf{\textbf{VNS}}}}}                                                                                                                                                                                                                              & \multicolumn{3}{c}{\textbf{MH}}  \\ \midrule
\textbf{Class}                                                                            & \textbf{N} & \textbf{Coloring} & \textbf{Total} & $\mathbf{\Delta_{Avg}^{Ini}}$ & $\mathbf{\Delta_{Avg}^{Fin}}$ & $\mathbf{\Delta_{Max}^{Fin}}$ & \textbf{t (s)} & $\mathbf{\Delta_{Avg}^{Fin}}$ & $\mathbf{\Delta_{Max}^{Fin}}$ & \textbf{t (s)} \\ 
\midrule
\multirow{8}{*}{\begin{tabular}[c]{@{}c@{}} \\ Random \\ 10 to 80 \end{tabular}}  & \multirow{2}{*}{102}   & Q2       & 10    & 3.90   & 2.90          & 6         & \textbf{54.00} & \textbf{2.50} & 6         & \textbf{54.00} \\
&                        & Q5 / Q20 & 20    & 1.75   & \textbf{1.20} & 5         & 30.12          & \textbf{1.20} & 5         & \textbf{30.10} \\ \cmidrule{2-11}
& \multirow{2}{*}{501}   & Q2       & 10    & 4.00   & 3.50          & 6         & 60.00          & \textbf{2.00} & 5         & \textbf{55.11} \\
&                        & Q5 / Q20 & 20    & 2.60   & 1.90          & 3         & 57.14          & \textbf{0.55} & 1         & \textbf{35.29} \\\cmidrule{2-11}
& \multirow{2}{*}{2001}  & Q2       & 10    & 5.10   & 4.40          & 7         & 60.00          & \textbf{0.80} & 2         & \textbf{43.59} \\
&                        & Q5 / Q20 & 20    & 5.90   & 3.90          & 8         & 60.00          & \textbf{0.15} & 1         & \textbf{15.22} \\ \cmidrule{2-11}
& \multirow{2}{*}{10002} & Q2       & 10    & 3.90   & 2.30          & 4         & 60.00          & \textbf{1.10} & 2         & \textbf{38.56} \\
&                        & Q5 / Q20 & 20    & 15.50  & 5.30          & 29        & 50.94          & \textbf{0.90} & 2         & \textbf{46.27} \\ \midrule
\multirow{16}{*}{\begin{tabular}[c]{@{}c@{}} \\ Random \\ 25 to 50 \end{tabular}} & \multirow{4}{*}{102}   & Q2       & 10    & 1.10   & \textbf{0.00} & 0         & \textbf{0.15}  & \textbf{0.00} & 0         & 0.20           \\
&                        & Q2H      & 10    & 1.30   & \textbf{0.00} & 0         & \textbf{0.09}  & \textbf{0.00} & 0         & 0.11           \\
&                        & Q5 / Q20 & 20    & 1.00   & \textbf{0.00} & 0         & \textbf{0.06}  & \textbf{0.00} & 0         & 0.16           \\
&                        & QN       & 10    & 1.30   & \textbf{0.00} & 0         & \textbf{0.08}  & \textbf{0.00} & 0         & 0.12           \\ \cmidrule{2-11}
& \multirow{4}{*}{501}   & Q2       & 10    & 1.10   & \textbf{0.00} & 0         & 0.37           & \textbf{0.00} & 0         & \textbf{0.36}  \\ 
&                        & Q2H      & 10    & 2.60   & \textbf{0.00} & 0         & \textbf{0.32}  & \textbf{0.00} & 0         & 0.62           \\
&                        & Q5 / Q20 & 20    & 1.05   & \textbf{0.00} & 0         & \textbf{0.49}  & \textbf{0.00} & 0         & 0.62           \\ 
&                        & QN       & 10    & 7.20   & \textbf{0.00} & 0         & \textbf{0.45}  & \textbf{0.00} & 0         & 0.51           \\ \cmidrule{2-11}
& \multirow{4}{*}{2001}  & Q2       & 10    & 2.70   & \textbf{0.50} & 1         & \textbf{31.03} & \textbf{0.50} & 1         & 31.76          \\ 
&                        & Q2H      & 10    & 16.00  & \textbf{0.50} & 1         & \textbf{32.15} & 0.80          & 1         & 49.90          \\
&                        & Q5 / Q20 & 20    & 2.05   & \textbf{0.50} & 1         & \textbf{33.52} & \textbf{0.50} & 1         & 35.86          \\
&                        & QN       & 10    & 30.90  & \textbf{0.50} & 1         & 36.62          & \textbf{0.50} & 1         & \textbf{33.35} \\  \cmidrule{2-11}
& \multirow{4}{*}{10002} & Q2       & 10    & 8.60   & 1.40          & 3         & \textbf{45.51} & \textbf{1.20} & 3         & 60.00          \\
&                        & Q2H      & 10    & 82.10  & \textbf{1.00} & 1         & \textbf{60.00} & \textbf{1.00} & 1         & \textbf{60.00} \\
&                        & Q5 / Q20 & 20    & 15.10  & \textbf{1.45} & 3         & \textbf{60.00} & 1.50          & 3         & \textbf{60.00} \\
&                        & QN       & 10    & 76.70  & \textbf{1.00} & 1         & \textbf{60.00} & 1.10          & 2         & \textbf{60.00} \\
\bottomrule
\end{tabular}
}
\caption{Results for the Random Instances.}
\label{tab::random}
\end{table}

Additionally, our approaches successfully handle all FK instances. Interestingly, instances with 120 items and Q2H coloring prove to be the most time-consuming for both approaches in this class.

Furthermore, it is important to highlight that Q2H coloring generally shows a larger~\(\Delta_{Avg}^{Ini}\) across all instance classes compared to Q2, Q5, and Q20 colorings. The only exception is the FK instances with 60 items. However, this does not necessarily imply that these instances are more challenging, as in some cases, the final maximum difference is slightly smaller for this coloring. 

Furthermore, the QN coloring has the largest~\(\Delta_{Avg}^{Ini}\) in the AI and Random25to50 classes, which could indicate that the Two-by-Two heuristic performs poorly specifically in BPP instances or instances with many colors. Nonetheless, this is not entirely the case, as running the BFD heuristic on these instances resulted in a~\(\Delta_{Avg}^{Ini}\) more than twice as large as that obtained with the Two-by-Two heuristic. 

Therefore, in colorings that impose a relatively restricted feasible solution space and are not as difficult to deal with as the Q2, Q5, and Q20 colorings, the Two-by-Two heuristic leverages the information provided by the color constraints to produce better results. On the other hand, if the coloring is highly challenging, such as Q2H coloring, or does not provide useful information, such as QN coloring, the Two-by-Two heuristic still outperforms a simple and effective BPP heuristic as the BFD\@.

Additionally, our MH algorithm outperforms VNS in terms of~\(\Delta_{Avg}^{Fin}\). MH optimally solves 210, 129, and 19 out of the 250 AI instances with 202, 403, and 601 items, respectively, whereas VNS fails to solve any of them. In some cases, the average final difference in Random10to80 instances is significantly smaller with MH than with VNS, with an average smaller than 3. Since the~$\Delta_{Avg}^{Fin}$ is larger in the Random10to80 class, this suggests that the lower bound~\(L_1\) may not be strong in this class, a conclusion supported by the results for small instances in the following section. 

Interestingly, VNS has smaller average times in some cases, although the discrepancy is not too significant in general. Particularly, VNS found 3 optimal solutions for the Random25to50 instances with 10002 items and Q2 coloring, while MH leaves a gap of at least 1 in all of these instances.

Finally, instances with QN coloring present times similar to Q5 and Q20 colorings, indicating that despite the poorer performance of the Two-by-Two heuristic, these BPP instances present similar difficulty for both our algorithms. Additionally, this contradicts the intuitive notion that the more restricted feasible solution space provided by color constraints makes the instance easier, as this is not the case for our algorithms in particular.

\subsection{Comparison with Literature's Algorithms}
Next, we compare our MH with five exact algorithms studied by~\cite{Borges_2023} named Basic Model~(BM), Color-alternating~(CA), Multilayered~(ML), Colored-clusters~(CC), and Color-resources~(CR). We obtained access to the source codes of these algorithms and ran them using the system configurations and Gurobi's version as presented in Section~\ref{section:Results}. Algorithms CC and CR, which rely on VRPSolver, were executed using the CPLEX solver version 20.1, while the others utilized the Gurobi solver.

As we propose a heuristic approach in this article, we impose a time limit of 60 seconds on all algorithms for comparison purposes. The implementations of BM, CA, and ML by the authors do not account for the time taken to build the mathematical model within the time limit. As we are working with large instances, constructing the mathematical model alone can take several minutes. To address this, we enforce a global time limit, terminating the algorithm's execution immediately after 120 seconds. For many large instances, these exact algorithms often do not produce any solution within this time limit. Therefore, we consider that the solution provided by each algorithm is the best between its solution and the BFD solution (where the BFD execution time is not included in the time limit).

\begin{table}[b!]
\resizebox{\columnwidth}{!}{%
\begin{tabular}{lrl|rr|rr|rrrrr}
\toprule
\textbf{Class}                                                                  & \textbf{\begin{tabular}[c]{@{}c@{}}Num.\\ Items\end{tabular}}      & \textbf{Coloring} & \textbf{Total} & $\mathbf{opt_{MH}}$ & $\mathbf{\Delta_{BFD}}$ & $\mathbf{\Delta_{MH}}$ & $\mathbf{\Delta_{KA}}$ & $\mathbf{\Delta_{CA}}$ & $\mathbf{\Delta_{MD}}$ & $\mathbf{\Delta_{CC}}$ & $\mathbf{\Delta_{CR}}$ \\ \midrule
\multirow{9}{*}{AI}                                                         & \multirow{3}{*}{202}   & Q2    & 50    & 33      & 9.2    & \textbf{0.2} & 1.3          & 9.2          & 9.2          & 6.3          & 5.8          \\
                                                                            &                        & Q2H   & 50    & 45      & 65.2   & \textbf{0.0} & 3.2          & 65.2         & 65.2         & 29.8         & 42.8         \\
                                                                            &                        & Q5 / Q20    & 100   & 87      & 1.3    & \textbf{0.0} & 0.9          & 1.3          & 1.3          & 1.2          & 1.3          \\ \cline{2-12}
                                                                            & \multirow{3}{*}{403}   & Q2    & 50    & 26      & 13.4   & \textbf{0.0} & 4.1          & 13.4         & 13.4         & 13.4         & 13.4         \\
                                                                            &                        & Q2H   & 50    & 31      & 132.5  & \textbf{0.0} & 10.4         & 132.5        & 132.5        & 132.5        & 132.5        \\
                                                                            &                        & Q5 / Q20    & 100   & 48      & 0.8    & \textbf{0.0} & 0.7          & 0.8          & 0.8          & 0.8          & 0.8          \\ \cline{2-12}
                                                                            & \multirow{3}{*}{601}   & Q2    & 50    & 3       & 15.0   & \textbf{0.0} & 8.4          & 15.0         & 15.0         & 15.0         & 15.0         \\
                                                                            &                        & Q2H   & 50    & 10      & 198.0  & \textbf{0.0} & 38.3         & 198.0        & 198.0        & 198.0        & 198.0        \\
                                                                            &                        & Q5 / Q20    & 100   & 5       & 0.4    & \textbf{0.0} & 0.3          & 0.4          & 0.4          & 0.4          & 0.4          \\ \midrule
\multirow{12}{*}{FK}                                                        & \multirow{3}{*}{60}    & Q2    & 20    & 20      & 5.8    & \textbf{0.0} & 0.9          & \textbf{0.0} & \textbf{0.0} & \textbf{0.0} & \textbf{0.0} \\
                                                                            &                        & Q2H   & 20    & 20      & 18.0   & \textbf{0.0} & 0.8          & \textbf{0.0} & \textbf{0.0} & \textbf{0.0} & \textbf{0.0} \\
                                                                            &                        & Q5 / Q20    & 40    & 40      & 3.3    & \textbf{0.0} & 1.0          & \textbf{0.0} & \textbf{0.0} & 0.0          & 1.6          \\ \cline{2-12}
                                                                            & \multirow{3}{*}{120}   & Q2    & 20    & 20      & 9.3    & \textbf{0.0} & 1.1          & \textbf{0.0} & \textbf{0.0} & \textbf{0.0} & \textbf{0.0} \\
                                                                            &                        & Q2H   & 20    & 20      & 35.5   & \textbf{0.0} & 2.9          & \textbf{0.0} & \textbf{0.0} & \textbf{0.0} & \textbf{0.0} \\
                                                                            &                        & Q5 / Q20    & 40    & 40      & 6.1    & \textbf{0.0} & 1.2          & \textbf{0.0} & 0.5          & 1.1          & 2.9          \\ \cline{2-12}
                                                                            & \multirow{3}{*}{249}   & Q2    & 20    & 20      & 21.9   & \textbf{0.0} & 2.2          & \textbf{0.0} & \textbf{0.0} & 17.5         & 0.2          \\
                                                                            &                        & Q2H   & 20    & 20      & 73.8   & \textbf{0.0} & 5.7          & \textbf{0.0} & \textbf{0.0} & 48.4         & 4.3          \\ 
                                                                            &                        & Q5 / Q20    & 40    & 40      & 12.3   & \textbf{0.0} & 2.6          & \textbf{0.0} & 6.1          & 9.0          & 6.2          \\ \cline{2-12}
                                                                            & \multirow{3}{*}{501}   & Q2    & 20    & 20      & 53.5   & \textbf{0.0} & 9.8          & 0.2          & \textbf{0.0} & 53.5         & 33.8         \\ 
                                                                            &                        & Q2H   & 20    & 20      & 151.2  & \textbf{0.0} & 33.5         & 43.5         & 18.6         & 121.5        & 151.2        \\
                                                                            &                        & Q5 / Q20    & 40    & 40      & 23.7   & \textbf{0.0} & 11.2         & 2.4          & 23.7         & 22.4         & 21.3         \\ \midrule
\multirow{8}{*}{\begin{tabular}[c]{@{}l@{}}Random\\ 10 to 80\end{tabular}}  & \multirow{2}{*}{102}   & Q2    & 10    & 10      & 3.4    & \textbf{0.0} & 0.2          & 1.5          & 1.2          & 0.9          & \textbf{0.0} \\
                                                                            &                        & Q5 / Q20    & 20    & 20      & 0.6    & \textbf{0.0} & \textbf{0.0} & 0.2          & 0.2          & \textbf{0.0} & 0.1          \\ \cline{2-12}
                                                                            & \multirow{2}{*}{501}   & Q2    & 10    & 6       & 12.0   & \textbf{0.0} & 4.1          & 5.2          & 5.2          & 12.0         & 10.3         \\
                                                                            &                        & Q5 / Q20    & 20    & 12      & 1.8    & \textbf{0.0} & 1.6          & 0.8          & 1.8          & 1.8          & 1.8          \\ \cline{2-12}
                                                                            & \multirow{2}{*}{2001}  & Q2    & 10    & 3       & 39.4   & \textbf{0.0} & 26.6         & 38.8         & 38.8         & 39.4         & 39.4         \\
                                                                            &                        & Q5 / Q20    & 20    & 17      & 4.3    & \textbf{0.0} & 4.2          & 4.3          & 4.3          & 4.3          & 4.3          \\ \cline{2-12}
                                                                            & \multirow{2}{*}{10002} & Q2    & 10    & 4       & 108.4  & \textbf{0.0} & 108.4        & 108.4        & 108.4        & 108.4        & 108.4        \\
                                                                            &                        & Q5 / Q20    & 20    & 7       & 14.1   & \textbf{0.0} & 14.1         & 14.1         & 14.1         & 14.1         & 14.1         \\ \midrule
\multirow{12}{*}{\begin{tabular}[c]{@{}l@{}}Random\\ 25 to 50\end{tabular}} & \multirow{3}{*}{102}   & Q2    & 10    & 10      & 8.9    & \textbf{0.0} & 1.0          & \textbf{0.0} & \textbf{0.0} & \textbf{0.0} & \textbf{0.0} \\
                                                                            &                        & Q2H   & 10    & 10      & 31.0   & \textbf{0.0} & 2.5          & \textbf{0.0} & \textbf{0.0} & \textbf{0.0} & \textbf{0.0} \\
                                                                            &                        & Q5 / Q20    & 20    & 20      & 4.8    & \textbf{0.0} & 1.2          & \textbf{0.0} & 2.5          & \textbf{0.0} & 2.4          \\ \cline{2-12}
                                                                            & \multirow{3}{*}{501}   & Q2    & 10    & 10      & 37.5   & \textbf{0.0} & 10.4         & 15.9         & 15.8         & 21.7         & 10.7         \\
                                                                            &                        & Q2H   & 10    & 10      & 150.4  & \textbf{0.0} & 35.9         & 102.1        & 79.7         & 135.2        & 135.3        \\
                                                                            &                        & Q5 / Q20    & 20    & 20      & 23.1   & \textbf{0.0} & 13.2         & 16.0         & 23.1         & 11.4         & 17.1         \\ \cline{2-12}
                                                                            & \multirow{3}{*}{2001}  & Q2    & 10    & 5       & 184.1  & \textbf{0.0} & 93.0         & 120.2        & 119.2        & 184.1        & 184.1        \\
                                                                            &                        & Q2H   & 10    & 2       & 589.1  & \textbf{0.0} & 528.9        & 589.1        & 589.1        & 589.1        & 589.1        \\
                                                                            &                        & Q5 / Q20    & 20    & 10      & 89.0   & \textbf{0.0} & 88.8         & 89.0         & 89.0         & 89.0         & 89.0         \\ \cline{2-12}
                                                                            & \multirow{3}{*}{10002} & Q2    & 10    & 0       & 1058.7 & \textbf{0.0} & 1058.7       & 570.8        & 570.8        & 1058.7       & 1058.7       \\ 
                                                                            &                        & Q2H   & 10    & 0       & 2960.6 & \textbf{0.0} & 2960.6       & 2960.6       & 2960.6       & 2960.6       & 2960.6       \\
                                                                            &                        & Q5 / Q20    & 20    & 0       & 445.1  & \textbf{0.0} & 445.1        & 445.1        & 445.1        & 445.1        & 445.1        \\ \bottomrule 

\end{tabular}
}
\caption{Comparison among MH and literature's algorithms.}
\label{tab::exacts}
\end{table}

In Table~\ref{tab::exacts}, we compare our MH, BFD, and the five exact algorithms of~\cite{Borges_2023}. The column~$\Delta_A$, where~$A$ represents an algorithm, displays the average difference between the solution of~$A$ and the best solution found by six algorithms across each instance class, number of items, and coloring. Since the dual bounds of these exact algorithms can excel the~$L_1$ bound, the column \textit{opt\textsubscript{MH}} indicates the number of optimal solutions found by the MH, considering the strongest bound found among all methods for each instance.

The table demonstrates that MH significantly outperforms the other algorithms, failing to find the best solution in only ten instances, all within the AI class with 202 items and Q2 coloring. Five of these best solutions were found by CR, and nine by CC (with some overlap). Additionally, in several cases, the averages reported by the exact algorithms match those of BFD\@. This suggests that either the exact algorithms failed to find any feasible solution or all solutions they found were worse than those of BFD\@.

Finally, it is noteworthy that the MH algorithm found an optimal solution for all instances with 102 items in the Random10to80 class, confirming the weakness of the L1 bound, as discussed earlier. Even for larger instances within this class, the MH algorithm managed to find numerous optimal solutions. However, concluding the quality of all solutions proves challenging. This is because there are instances, including those with 501 items, where only the BM algorithm was able to calculate the lower bound, which is known to give a bound equal to L1 in BPP instances.

\section{Conclusion}\label{section:Conclusion}

Our approaches have proven highly effective, enabling us to find solutions very close to optimal, even for instances with a large number of items. This success can be attributed in part to the high-quality solutions provided by Two-by-Two, which adeptly manages the color constraint. Another positive factor is our neighborhoods, which are efficient in quality and time complexity. 

Additionally, Kantorovich-Gilmore-Gomory Formulation has a strong relaxation for the BPP, and this property may extend to CBPP\@. A strong relaxation implies that values for the optimal fractional and integer solutions are very close, suggesting that the patterns used by the relaxation likely belong to high-quality integer solutions. Therefore, this relaxation solution proves to be an excellent method for generating a fruitful set of initial patterns for GRASP\@.

Moreover, CBPP does not appear to be an easier version of BPP, despite having a more restricted set of feasible solutions. In our algorithms, the performance on BPP instances is similar to that on instances with Q5 and Q20 colorings. This perception is supported by results obtained by~\citet{Borges_2023}, where both exact approaches using VRPSolver perform worse compared to the VRPSolver's performance on BPP instances.

In future work, it is interesting to use a better lower bound to evaluate Random10to80 instances since $L_1$ can be weak, and we may be closer to the optimal solution without knowing. However, this task is likely to be challenging. Notably, the exact algorithms examined by \cite{Borges_2023} encountered difficulties in solving their relaxations within our specified time constraints for several instances. This challenge likely stems from the complexity of our benchmarks, ranging from those with a high number of items~$(n = 10^4 + 2)$ to those with substantial bin capacity $(W = 10^6 +1)$. Furthermore, the relaxations of the BPP and its variants typically involve subroutines with a time complexity of~$\Omega(n \cdot W)$, a feature also shared by the exact CBPP algorithms mentioned here. Moreover, there is potential to extend our efficient algorithms to other BPP variants by adapting the dynamic programming approach of the Auxiliary Algorithm to accommodate additional constraints while maintaining linear time complexity.

\section*{Acknowledgements} This project was supported by the São Paulo Research Foundation \-(FAPESP) grants \#2020/06511-0 and \#2022/05803-3; and the Brazilian National Council For Scientific and Technological Development~(CNPq) grants \#144257/2019-0 and \#311039/2020-0.

\section*{Declarations}
\textbf{Conflict of interest} All authors declare that they have no conflicts of interest.


\begin{thebibliography}{37}
\providecommand{\natexlab}[1]{#1}
\providecommand{\url}[1]{{#1}}
\providecommand{\urlprefix}{URL }
\providecommand{\doi}[1]{\url{https://doi.org/#1}}
\providecommand{\eprint}[2][]{\url{#2}}
    \bibcommenthead

\bibitem[{Alsarhan et~al.(2016)Alsarhan, Chia, Christman, Fu, and
    Jin}]{Alsarhan_2016}
Alsarhan H, Chia D, Christman A, et~al. (2016) A two-pass algorithm for
    unordered colored bin packing. Proceedings of the 9th International
    Conference on Discrete Optimization and Operations Research and Scientific
    School pp 1--10

\bibitem[{Baldacci et~al.(2024)Baldacci, Coniglio, Cordeau, and
    Furini}]{Baldacci_2024}
Baldacci R, Coniglio S, Cordeau JF, et~al. (2024) A numerically exact algorithm
    for the bin-packing problem. INFORMS Journal on Computing 36(1):141--162.
    \doi{10.1287/ijoc.2022.0257}

\bibitem[{Balogh et~al.(2013)Balogh, B{\'e}k{\'e}si, Dosa, Kellerer, and
    Tuza}]{Balogh_2012}
Balogh J, B{\'e}k{\'e}si J, Dosa G, et~al. (2013) Black and white bin packing.
    In: Erlebach T, Persiano G (eds) Approximation and Online Algorithms.
    Springer Berlin Heidelberg, Berlin, Heidelberg, pp 131--144,
    \doi{10.1007/978-3-642-38016-7_12}

\bibitem[{Balogh et~al.(2015{\natexlab{a}})Balogh, B{\'e}k{\'e}si, D{\'o}sa,
    Epstein, Kellerer, and Tuza}]{Balogh_2015b}
Balogh J, B{\'e}k{\'e}si J, D{\'o}sa G, et~al. (2015{\natexlab{a}}) Online
    results for black and white bin packing. Theory of Computing Systems
    56(1):137--155. \doi{10.1007/s00224-014-9538-8}

\bibitem[{Balogh et~al.(2015{\natexlab{b}})Balogh, Békési, Dósa, Epstein,
    Kellerer, Levin, and Tuza}]{Balogh_2015}
Balogh J, Békési J, Dósa G, et~al. (2015{\natexlab{b}}) Offline black and
    white bin packing. Theoretical Computer Science 596:92--101.
    \doi{10.1016/j.tcs.2015.06.045}

\bibitem[{Belov and Scheithauer(2006)}]{Belov_2006}
Belov G, Scheithauer G (2006) A branch-and-cut-and-price algorithm for
    one-dimensional stock cutting and two-dimensional two-stage cutting. European
    Journal of Operational Research 171(1):85--106.
    \doi{10.1016/j.ejor.2004.08.036}

\bibitem[{B{\"o}hm et~al.(2015)B{\"o}hm, Sgall, and Vesel{\'y}}]{Bohm_2014}
B{\"o}hm M, Sgall J, Vesel{\'y} P (2015) Online colored bin packing. In: Bampis
    E, Svensson O (eds) Approximation and Online Algorithms. Springer
    International Publishing, Cham, pp 35--46, \doi{10.1007/978-3-319-18263-6_4}

\bibitem[{B{\"o}hm et~al.(2018)B{\"o}hm, D{\'o}sa, Epstein, Sgall, and
    Vesel{\'y}}]{Bohm_2018}
B{\"o}hm M, D{\'o}sa G, Epstein L, et~al. (2018) Colored bin packing: Online
    algorithms and lower bounds. Algorithmica 80(1):155--184.
    \doi{10.1007/s00453-016-0248-2}

\bibitem[{Borges et~al.(2024)Borges, Schouery, and Miyazawa}]{Borges_2023}
Borges YG, Schouery RC, Miyazawa FK (2024) Mathematical models and exact
    algorithms for the colored bin packing problem. Computers \& Operations
    Research p 106527. \doi{10.1016/j.cor.2023.106527}

\bibitem[{Brandão and Pedroso(2016)}]{Brandao_2016}
Brandão F, Pedroso JP (2016) Bin packing and related problems: General
    arc-flow formulation with graph compression. Computers \& Operations Research
    69:56--67. \doi{10.1016/j.cor.2015.11.009}

\bibitem[{Buljubašić and Vasquez(2016)}]{Buljubasic_2016}
Buljubašić M, Vasquez M (2016) Consistent neighborhood search for
    one-dimensional bin packing and two-dimensional vector packing. Computers \&
    Operations Research 76:12--21. \doi{10.1016/j.cor.2016.06.009}

\bibitem[{Carvalho(1999)}]{Carvalho_1999}
Carvalho JMVd (1999) Exact solution of bin‐packing problems using column
    generation and branch‐and‐bound. Annals of Operations Research
    86(0):629--659. \doi{10.1023/A:1018952112615}

\bibitem[{Castelli and Vanneschi(2014)}]{Castelli_2014}
Castelli M, Vanneschi L (2014) A hybrid harmony search algorithm with variable
    neighbourhood search for the bin-packing problem. 2014 Sixth World Congress
    on Nature and Biologically Inspired Computing (NaBIC 2014) pp 1--6.
    \doi{10.1109/NaBIC.2014.6921849}

\bibitem[{Chen et~al.(2015)Chen, Han, Bein, and Ting}]{Chen_2015}
Chen J, Han X, Bein W, et~al. (2015) Black and white bin packing revisited. In:
    Lu Z, Kim D, Wu W, et~al. (eds) Combinatorial Optimization and Applications.
    Springer International Publishing, Cham, pp 45--59,
    \doi{10.1007/978-3-319-26626-8_4}

\bibitem[{Delorme and Iori(2020)}]{Delorme_2020}
Delorme M, Iori M (2020) Enhanced pseudo-polynomial formulations for bin
    packing and cutting stock problems. INFORMS Journal on Computing
    32(1):101--119. \doi{10.1287/ijoc.2018.0880}

\bibitem[{Delorme et~al.(2018)Delorme, Iori, and Martello}]{Delorme_2018}
Delorme M, Iori M, Martello S (2018) Bpplib: a library for bin packing and
    cutting stock problems. Optimization Letters 12(2):235--250.
    \doi{10.1007/s11590-017-1192-z}

\bibitem[{Dem{\v{s}}ar(2006)}]{Demvsar_2006}
Dem{\v{s}}ar J (2006) Statistical comparisons of classifiers over multiple data
    sets. The Journal of Machine learning research 7:1--30

\bibitem[{D{\'o}sa and Epstein(2014)}]{Dosa_2014}
D{\'o}sa G, Epstein L (2014) Colorful bin packing. In: Ravi R, G{\o}rtz IL
    (eds) Algorithm Theory -- SWAT 2014. Springer International Publishing, Cham,
    pp 170--181, \doi{10.1007/978-3-319-08404-6_15}

\bibitem[{Feo and Resende(1989)}]{Resende_1989}
Feo TA, Resende MG (1989) A probabilistic heuristic for a computationally
    difficult set covering problem. Operations Research Letters 8(2):67--71.
    \doi{https://doi.org/10.1016/0167-6377(89)90002-3}

\bibitem[{Fleszar and Hindi(2002)}]{Fleszar_2002}
Fleszar K, Hindi KS (2002) New heuristics for one-dimensional bin-packing.
    Computers \& Operations Research 29(7):821--839.
    \doi{10.1016/S0305-0548(00)00082-4}

\bibitem[{Friedman(1937)}]{Friedman_1937}
Friedman M (1937) The use of ranks to avoid the assumption of normality
    implicit in the analysis of variance. Journal of the American Statistical
    Association 32(200):675--701. \doi{10.1080/01621459.1937.10503522}

\bibitem[{Gilmore and Gomory(1961)}]{Gilmore_1961}
Gilmore PC, Gomory RE (1961) A linear programming approach to the cutting-stock
    problem. Operations Research 9(6):849--859. \doi{10.1287/opre.29.6.1092}

\bibitem[{Gonz{\'a}lez-San-Mart{\'i}n et~al.(2023)Gonz{\'a}lez-San-Mart{\'i}n,
    Cruz-Reyes, G{\'o}mez-Santill{\'a}n, Fraire, Rangel-Valdez, Dorronsoro, and
    Quiroz-Castellanos}]{Gonzalez_2023}
Gonz{\'a}lez-San-Mart{\'i}n J, Cruz-Reyes L, G{\'o}mez-Santill{\'a}n C, et~al.
    (2023) Comparative Study of Heuristics for the One-Dimensional Bin Packing
    Problem, Springer Nature Switzerland, Cham, pp 293--305.
    \doi{10.1007/978-3-031-28999-6_19}

\bibitem[{Gupta and Ho(1999)}]{Gupta_1999}
Gupta JND, Ho JC (1999) A new heuristic algorithm for the one-dimensional
    bin-packing problem. Production Planning \& Control 10(6):598--603.
    \doi{10.1080/095372899232894}

\bibitem[{Iman and Davenport(1980)}]{Iman_1980}
Iman RL, Davenport JM (1980) Approximations of the critical region of the
    fbietkan statistic. Communications in Statistics - Theory and Methods
    9(6):571--595. \doi{10.1080/03610928008827904}

\bibitem[{Kantorovich(1939)}]{Kantorovich1939}
Kantorovich L (1939) Matematicheskie metody organizatsii i planirovaniya
    proizvodstva [\textit{Mathematical methods of organizing and planning
    production}]. Lenizdat, Leningrad

\bibitem[{Kantorovich and Zalgaller(1951)}]{Kantorovich1951}
Kantorovich L, Zalgaller V (1951) Ratsionalnyj raskroj promyshlennykh
    materialov [\textit{Calculation of rational cutting of stock}]. Lenizdat,
    Leningrad

\bibitem[{Karp(1972)}]{Karp_1972}
Karp RM (1972) Reducibility among Combinatorial Problems, Springer US, Boston,
    MA, pp 85--103. \doi{10.1007/978-1-4684-2001-2_9}

\bibitem[{de~Lima et~al.(2022)de~Lima, Iori, and Miyazawa}]{Loti_2022}
de~Lima VL, Iori M, Miyazawa FK (2022) Exact solution of network flow models
    with strong relaxations. Mathematical Programming pp 1--34.
    \doi{10.1007/s10107-022-01785-9}

\bibitem[{Loh et~al.(2008)Loh, Golden, and Wasil}]{Loh_2008}
Loh KH, Golden B, Wasil E (2008) Solving the one-dimensional bin packing
    problem with a weight annealing heuristic. Computers \& Operations Research
    35(7):2283--2291. \doi{10.1016/j.cor.2006.10.021}

\bibitem[{Martello and Toth(1990)}]{Martello1990}
Martello S, Toth P (1990) Knapsack problems: algorithms and computer
    implementations. John Wiley \& Sons, Inc.

\bibitem[{Mladenović and Hansen(1997)}]{Mladenovic_1997}
Mladenović N, Hansen P (1997) Variable neighborhood search. Computers \&
    Operations Research 24(11):1097--1100. \doi{10.1016/S0305-0548(97)00031-2}

\bibitem[{Nemenyi(1963)}]{Nemenyi_1963}
Nemenyi PB (1963) Distribution-free multiple comparisons. PhD thesis, Princeton
    University

\bibitem[{Pessoa et~al.(2020)Pessoa, Sadykov, Uchoa, and
    Vanderbeck}]{Pessoa_2020}
Pessoa A, Sadykov R, Uchoa E, et~al. (2020) A generic exact solver for vehicle
    routing and related problems. Mathematical Programming 183(1):483--523.
    \doi{10.1007/s10107-020-01523-z}

\bibitem[{Uchoa and Sadykov(2024)}]{Uchoa2024}
Uchoa E, Sadykov R (2024) {Kantorovich and Zalgaller} (1951): the 0-th column
    generation algorithm. Tech. Rep. L-2024-1, Cadernos do LOGIS-UFF,
    Niter{\'o}i, Brazil

\bibitem[{Vance(1998)}]{Vance_1998}
Vance PH (1998) Branch-and-price algorithms for the one-dimensional cutting
    stock problem. Computational Optimization and Applications 9(3):211--228.
    \doi{10.1023/A:1018346107246}

\bibitem[{Wei et~al.(2020)Wei, Luo, Baldacci, and Lim}]{Wei_2019}
Wei L, Luo Z, Baldacci R, et~al. (2020) A new branch-and-price-and-cut
    algorithm for one-dimensional bin-packing problems. INFORMS Journal on
    Computing 32(2):428--443. \doi{10.1287/ijoc.2018.0867}

\end{thebibliography}
\end{document}